\documentclass[journal]{IEEEtran}
\usepackage{times}
\usepackage{epsfig}
\usepackage{graphicx}
\usepackage{amsmath}
\usepackage{amssymb}
\usepackage{ctable}
\usepackage{multirow}
\usepackage{hhline}
\usepackage{stfloats}
\usepackage{colortbl}		
\usepackage{epsfig}
\usepackage{graphicx}
\usepackage{array}
\usepackage{amsmath}
\usepackage{array}
\usepackage{hhline}
\usepackage{colortbl}
\usepackage{amssymb}
\usepackage{stfloats}
\usepackage{multirow}
\usepackage{booktabs}
\usepackage{threeparttable}
\usepackage{cite}
\usepackage{float}
\usepackage{eso-pic}
\usepackage{subfig}

\usepackage{multirow, booktabs}
\ifCLASSINFOpdf
\else
\fi

\usepackage[pagebackref=true,breaklinks=true,letterpaper=true,colorlinks,bookmarks=false]{hyperref}
\definecolor{mygray}{gray}{.9}
\newcommand{\red}[1]{\textcolor{red}{#1}}

\definecolor{amber}{rgb}{1.0, 0.49, 0.0}

\begin{document}

\title{Unifying Global-Local Representations in \\Salient Object Detection with Transformers}

\author{Sucheng Ren,
       Nanxuan Zhao,
       Qiang Wen,
       Guoqiang Han,
       and Shengfeng He, \IEEEmembership{Senior Member, IEEE}
\thanks{The work is supported by the Guangdong Natural Science Funds for Distinguished Young Scholars (No. 2023B1515020097), Singapore MOE Tier 1 Funds (MSS23C002), and the National Research Foundation Singapore under the AI Singapore Programme (No: AISG3-GV-2023-011). (Shengfeng He is the corresponding author.)}
\thanks{Sucheng Ren and Shengfeng He are with the School of Computing and Information Systems, Singapore Management University, Singapore. E-mail: oliverrensu@gmail.com, shengfenghe@smu.edu.sg.}
\thanks{Nanxuan Zhao is with Adobe Research, San Jose, USA. E-mail: nanxuanzhao@gmail.com.}
\thanks{Qiang Wen and Guoqiang Han are with the School of Computer Science and Engineering, South China University of Technology, Guangzhou, China. E-mail: csqiangwen@gmail.com, csgqhan@scut.edu.cn.}
\thanks{ @2024 IEEE.  Personal use of this material is permitted.  Permission from IEEE must be obtained for all other uses, in any current or future media, including reprinting/republishing this material for advertising or promotional purposes, creating new collective works, for resale or redistribution to servers or lists, or reuse of any copyrighted component of this work in other works.}}

\markboth{IEEE Transactions on Emerging Topics in Computational Intelligence}
{Shell \MakeLowercase{\textit{Ren et al.}}: Unifying Global-Local Representations in Salient Object Detection with Transformers}

\maketitle

\def\etal{\textit{et al.~}}
\def\ie{\textit{i.e.~}}
\def\eg{\textit{e.g.~}}

\begin{abstract}
	The fully convolutional network (FCN) has dominated salient object detection for a long period. However, the locality of CNN requires the model deep enough to have a global receptive field and such a deep model always leads to the loss of local details. In this paper, we introduce a new attention-based encoder, vision transformer, into salient object detection to ensure the globalization of the representations from shallow to deep layers. With the global view in very shallow layers, the transformer encoder preserves more local representations to recover the spatial details in final saliency maps. Besides, as each layer can capture a global view of its previous layer, adjacent layers can implicitly maximize the representation differences and minimize the redundant features, making every output feature of transformer layers contribute uniquely to the final prediction. To decode features from the transformer, we propose a simple yet effective deeply-transformed decoder. The decoder densely decodes and upsamples the transformer features, generating the final saliency map with less noise injection. Experimental results demonstrate that our method significantly outperforms other FCN-based and transformer-based methods in five benchmarks by a large margin, with an average of 12.17\% improvement in terms of Mean Absolute Error (MAE). Code is available at \href{https://github.com/OliverRensu/GLSTR}{https://github.com/OliverRensu/GLSTR}.
\end{abstract}

\begin{IEEEkeywords}
Transformer, Salient object detection.
\end{IEEEkeywords}

\IEEEpeerreviewmaketitle

\section{Introduction}
\begin{figure}[t]
	\centering
	\setlength{\tabcolsep}{.5pt}
	\renewcommand{\arraystretch}{.5}
	\begin{tabular}{cccc}
		\includegraphics[width=.245\linewidth]{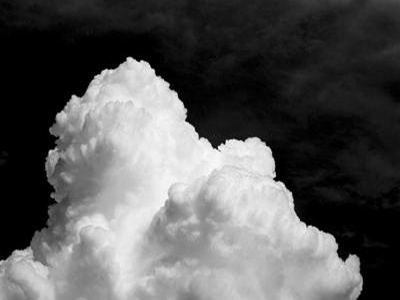}&
		\includegraphics[width=.245\linewidth]{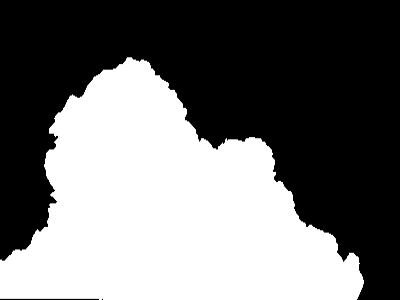}&
		\includegraphics[width=.245\linewidth]{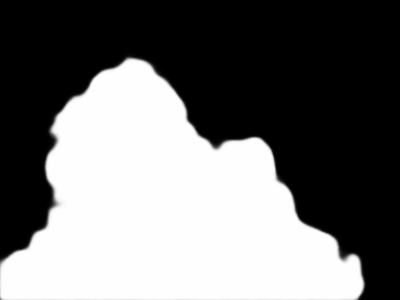}&
		\includegraphics[width=.245\linewidth]{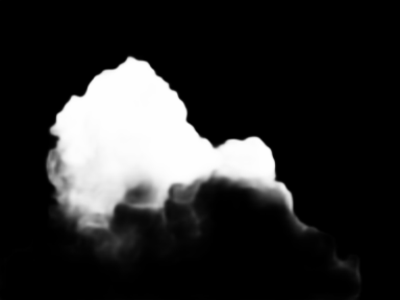}\\

		\includegraphics[width=.245\linewidth]{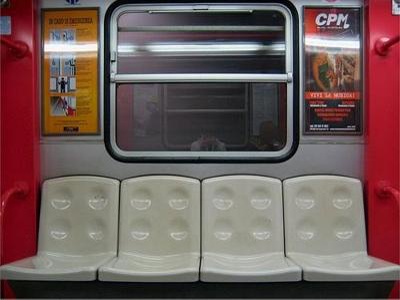}&
		\includegraphics[width=.245\linewidth]{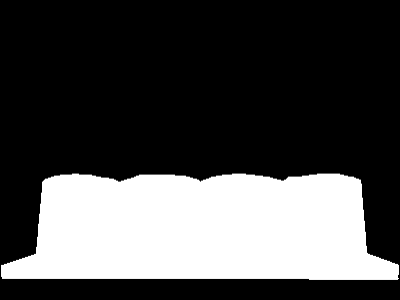}&
		\includegraphics[width=.245\linewidth]{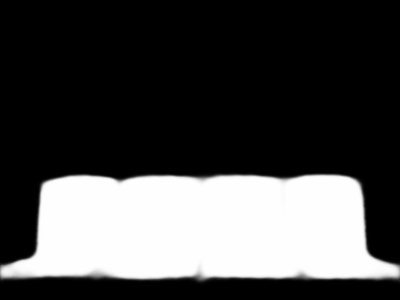}&
		\includegraphics[width=.245\linewidth]{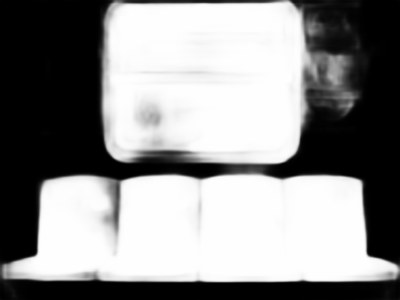}\\

		\includegraphics[width=.245\linewidth]{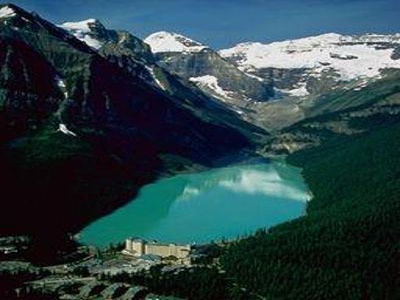}&
		\includegraphics[width=.245\linewidth]{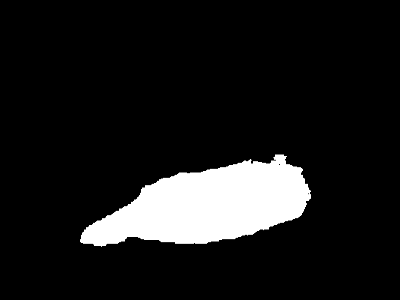}&
		\includegraphics[width=.245\linewidth]{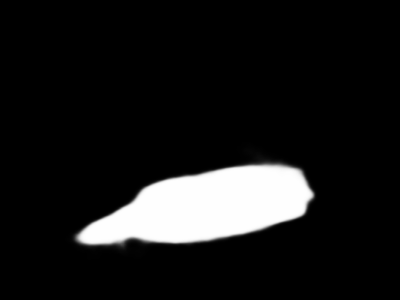}&
		\includegraphics[width=.245\linewidth]{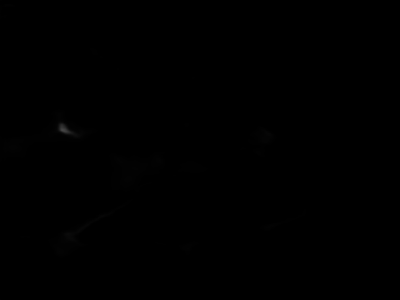}\\

		\includegraphics[width=.245\linewidth]{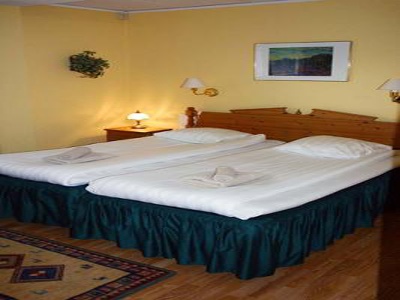}&
		\includegraphics[width=.245\linewidth]{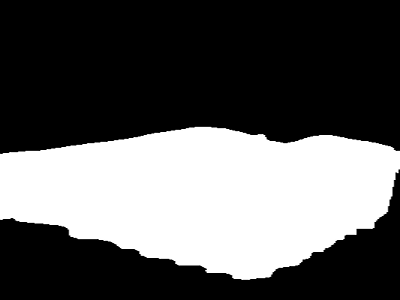}&
		\includegraphics[width=.245\linewidth]{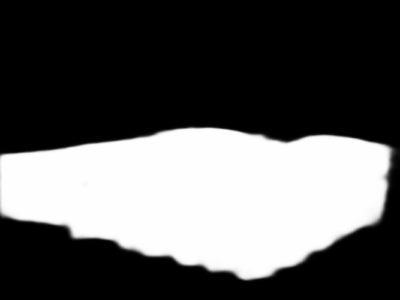}&
		\includegraphics[width=.245\linewidth]{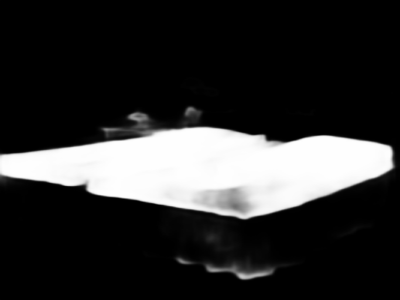}\\
	
		Input & GT & Ours & GateNet
	\end{tabular}
	\caption{Examples of our method. We propose the Global-Local Saliency TRansformer (GLSTR) to unify global and local features in each layer. We compare our method with a SOTA method, GateNet~\cite{zhao2020suppress}, based on FCN architecture. Our method can localize salient region precisely with accurate boundary. }
	\label{img:teaser}
\end{figure}

Salient object detection (SOD) aims at segmenting the most visually attracting objects of the images, aligning with the human perception. It gained broad attention in recent years due to its fundamental role in many vision tasks, such as image manipulation~\cite{mechrez2019saliency}, segmentation~\cite{luo2021weakly,wang2022looking,wang2017saliency}, autonomous navigation~\cite{craye2016environment}, person re-identification~\cite{yang2014salient,zhao2016person}, and photo cropping~\cite{wang2018deep}.

Traditional methods~\cite{cheng2014global,r1_1,cheng2014global,einhauser2003does,he2014saliency,r1_4} mainly rely on different priors and hand-crafted features like color contrast~\cite{cheng2014global}, brightness~\cite{einhauser2003does}, foreground and background distinctness~\cite{he2014saliency}. However, these methods limit the representation power, without considering semantic information. Thus, they often fail to deal with complex scenes. With the development of deep convolutional neural network~(CNN), fully convolutional network~(FCN)~\cite{long2015fully} becomes an essential building block for SOD~\cite{li2016deep, wang2017stagewise, wang2016saliency, zhang2018bi}. These methods can encode high-level semantic features and better localize the salient regions.

However, due to the nature of locality, methods relying on FCN usually face a trade-off between capturing global and local features. To encode high-level global representations, the model needs to take a stack of convolutional layers to obtain larger receptive fields, but it erases local details. To retain the local understanding of saliency, features from shallow layers fail to incorporate high-level semantics. This discrepancy may make the fusion of global and local features less efficient. Is it possible to unify global and local features in each layer to reduce the ambiguity and enhance the accuracy for SOD?

\begin{figure}[t]
	\centering
	\setlength{\tabcolsep}{.5pt}
	\renewcommand{\arraystretch}{.5}
	\begin{tabular}{cccc}
		\includegraphics[width=.245\linewidth]{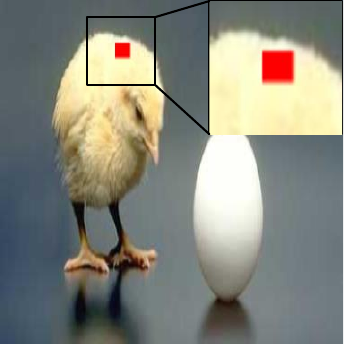}&
		\includegraphics[width=.245\linewidth]{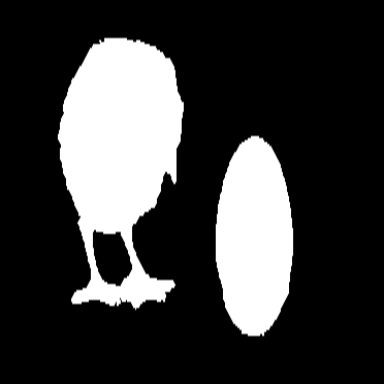}&
		\includegraphics[width=.245\linewidth]{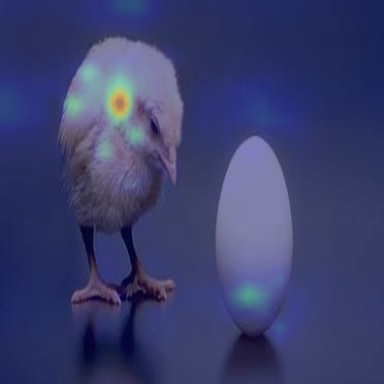}&
		\includegraphics[width=.245\linewidth]{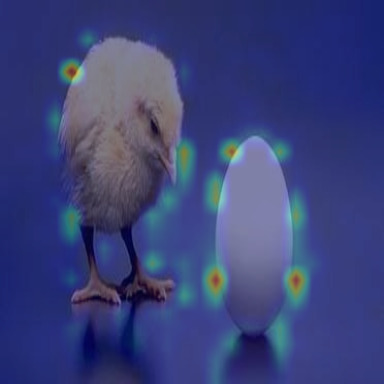}\\
		
		\includegraphics[width=.245\linewidth]{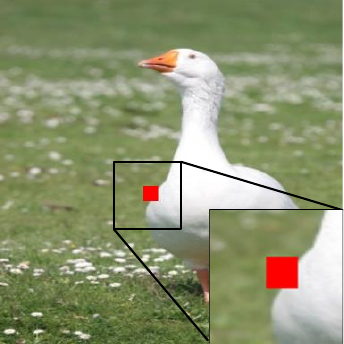}&
		\includegraphics[width=.245\linewidth]{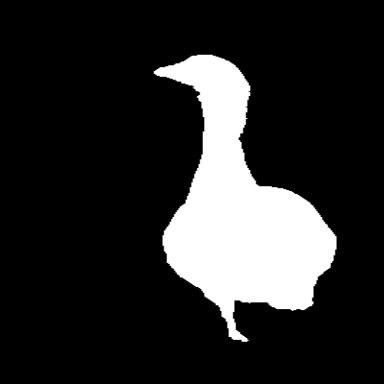}&
		\includegraphics[width=.245\linewidth]{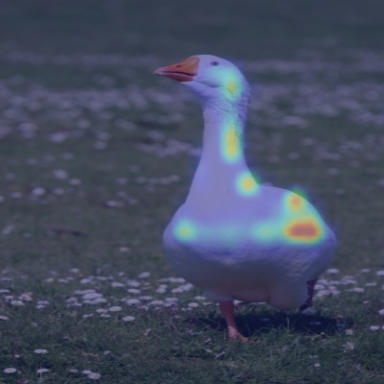}&
		\includegraphics[width=.245\linewidth]{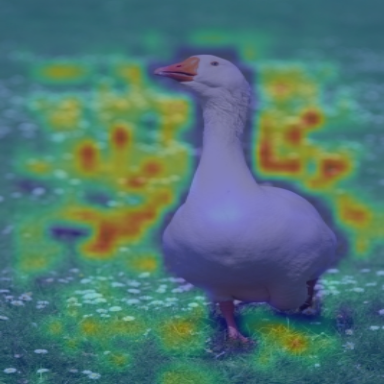}\\
		Input & GT & Layer 1 & Layer 12
	\end{tabular}
	
	\caption{The visual attention map of the red block in the input image on the first and twelfth layers of the transformer. The attention maps show that features from shallow layer can also have global information and features from deep layer can also have local information.}
	\label{img:attention}
\end{figure}

In this paper, we aim to jointly learn global and local features in a layer-wise manner for solving the salient object detection task. Rather than using a pure CNN architecture, we turn to transformer~\cite{vaswani2017attention} for help. We get inspired from the recent success of transformer on various vision tasks~\cite{dosovitskiy2020image} for its superiority in exploring long-range dependency. Transformer applies self-attention mechanism for each layer to learn a global representation. Therefore, it is possible to inject globality in the shallow layers, while maintaining local features. As illustrated in Fig.~\ref{img:attention}, the attention map of the red block has the global view of the chicken and the egg even in the first layer. The network does attend to small scale details even in the final layer (the twelfth layer). More importantly, with the self-attention mechanism, the transformer is capable to model the ``contrast'', which has demonstrated to be crucial for saliency perception~\cite{reynolds2003interacting, morgan1943physiological, desimone1995neural}.

With the merits mentioned above, we propose a method called \textbf{G}lobal-\textbf{L}ocal \textbf{S}aliency \textbf{TR}ansformer (\textbf{GLSTR}). The core of our method is a pure transformer-based encoder and a mixed decoder to aggregate features generated by transformers. To encode features through transformers, we first split the input image into a grid of fixed-size patches. We use a linear projection layer to map each image patch to a feature vector for representing local details. By passing the feature through the multi-head self-attention module in each transformer layer, the model further encodes global features without diluting the local ones. To decode the features with global-local information over the inputs and the previous layer from the transformer encoder, we propose to densely decode feature from each transformer layer. With this dense connection, the representations during decoding still preserve rich local and global features. We propose a novel framework to apply the pure transformer-based encoder on SOD tasks.

To sum up, our contributions are in three folds:
\begin{itemize}
	\item We unify global-local representations with a novel transformer-based architecture, which models long-range dependency within each layer.
	\item To take full advantage of the global information of previous layers, we propose a new decoder, deeply-transformed decoder, to densely decode features of each layer.
	\item We conduct extensive evaluations on five widely-used benchmark datasets, showing that our method outperforms state-of-the-art methods by a large margin, with an average of $12.17\%$ improvement in terms of MAE.
\end{itemize}

\section{Related Work}
    \textbf{Salient Object Detection.}
    FCN has become the mainstream architecture for the saliency detection methods, to predict pixel-wise saliency maps directly \cite{li2016deep, wang2017stagewise, wang2016saliency, zhang2018bi, liu2018picanet,10013775, zhang2018progressive, he2017delving, ren2020tenet,Ren_2021_CVPR, wang2020learning,r1_2,r1_3}. There are several surveys\cite{wang2021salient,borji2019salient} about FCN-based SOD methods. Especially, the skip-connection architecture has been demonstrated its performance in combining global and local context information \cite{hou2017deeply, zhao2020suppress, liu2019simple, zhang2017amulet, luo2017non, wang2018detect, simonyan2014very,Wang_2019_CVPR,wang2019inferring,xu2021locate}.
    Hou \etal \cite{hou2017deeply} proposed that the high-level features are capable of localizing the salient regions while the shallower ones are active at preserving low-level details. Therefore, they introduced short connections between the deeper and shallower feature pairs.
    Based on the U-Shape architecture, Zhao \etal \cite{zhao2020suppress} introduced a transition layer and a gate into every skip connection, to suppress the misleading convolutional information in the original features from the encoder. To localize the salient objects better, they also built a Fold-ASPP module on the top of the encoder.
    Similarly, Liu \etal \cite{liu2019simple} inserted a pyramid pooling module on the top of the encoder to capture the high-level semantic information. To recover the diluted information in the convolutional encoder and alleviate the coarse-level feature aggregation problem while decoding, they introduced a feature aggregation module after every skip connection.
    In \cite{zhao2019egnet}, Zhao \etal adopted the VGG network without fully connected layers as their backbone. They fused the high-level feature into the shallower one at each scale to recover the diluted location information. After three convolution layers, each fused feature was enhanced by the saliency map. Specifically, they introduced the edge map as guidance for the shallowest feature and fused this feature into deeper ones to produce the side outputs.

    The aforementioned methods usually suffer from the limited size of convolutional operation. Although there will be large receptive fields in the deep convolution layers, the low-level information is diluted increasingly with the increasing number of layers.

    \textbf{Transformer.}
    Since the local information covered by the convolutional operation is deficient, numerous methods adopt attention mechanisms to capture the long-range correlation \cite{wang2018non, bello2019attention, hu2019local, zhao2020exploring,ren2022shunted,ren2023tinymim,ren2023sg, wang2020axial}. The transformer, proposed by Vaswani \etal \cite{vaswani2017attention}, has demonstrated its global-attention performance in machine translation tasks. Recently, more and more works focus on exploring the ability of transformers in computer vision tasks.
    Dosovitskiy \etal \cite{dosovitskiy2020image} proposed to apply the pure transformer directly to sequences of image patches for exploring spatial correlation on image classification tasks.
    Carion \etal~\cite{carion2020end} proposed a transformer encoder-decoder architecture on object detection tasks. By collapsing the feature from CNN into a sequence of learned object queries, the transformer can dig out the correlation between objects and the global image context.
    Zeng \etal \cite{zeng2020learning} proposed a spatial-temporal transformer network to tackle video inpainting problems along spatial and temporal channels jointly.

    The most related work to ours is \cite{zheng2020rethinking}. They proposed to replace the convolution layers with the pure transformer during encoding for the semantic segmentation task. Unlike this work, we propose a transformer-based U-shape architecture network that is more suitable for dense prediction tasks. With our dense skip connections and progressive upsampling strategy, the proposed method can sufficiently associate the decoding features with the global-local ones from the transformer.

\begin{figure*}[t]
    \centering
    \includegraphics[width=\linewidth]{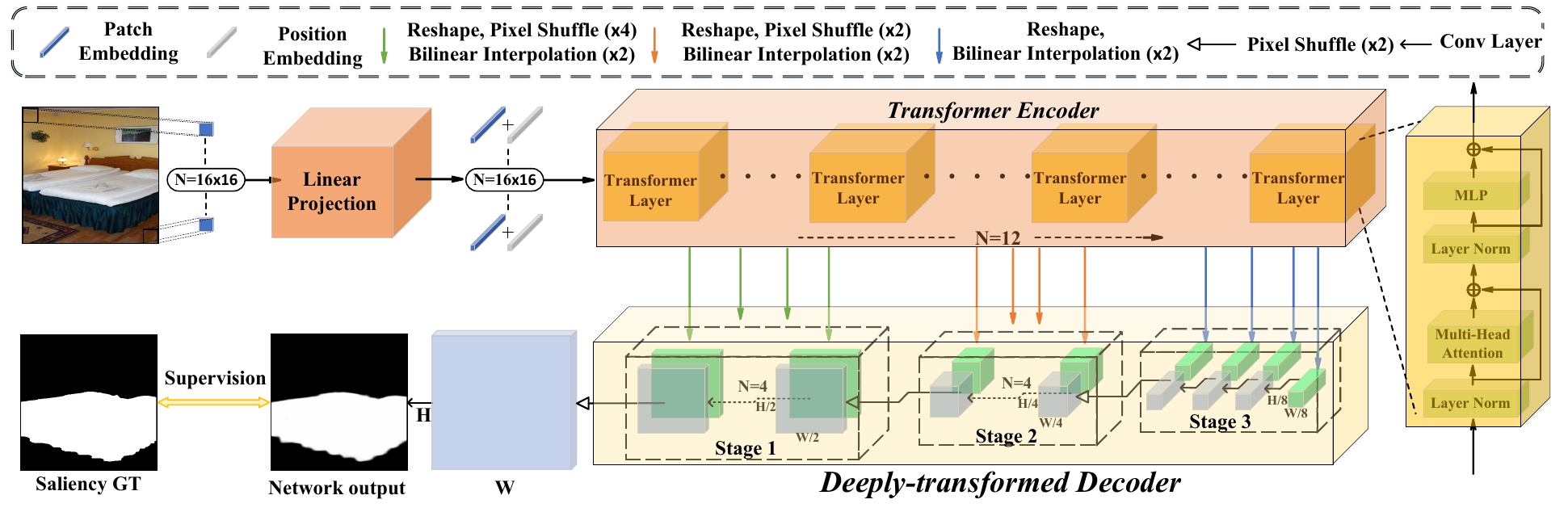}
    \caption{The pipeline of our proposed method. We first divide the image into non-overlap patches and map each patch into the token before feeding to transformer layers. After encoding features through 12 transformer layers, we decode each output feature in three successive stages with 8$\times$, 4$\times$, and 2$\times$ upsampling respectively. Each decoding stage contains four layers and the input of each layer comes from the features of its previous layer together with the corresponding transformer layer.}
    \label{fig:illustration}
\end{figure*}

\section{Method}
In this paper, we propose a Global-Local Saliency TRansfromer (GLSTR) which models the local and global representations jointly in each encoder layer and decodes all features from transformers with gradual spatial resolution recovery. In the following subsections, we first review the local and global feature extraction by the fully convolutional network-based salient object detection methods in Sec. \ref{sec:FCN}. Then we give more details about how the transformer extracts global and local features in each layer in Sec. \ref{sec:transformer}. Finally, we specify different decoders including naive decoder, stage-by-stage decoder, multi-level feature aggregation decoder~\cite{zheng2020rethinking}, and our proposed decoder. The framework of our proposed method is shown in Fig. \ref{fig:illustration}.

\subsection{FCN-based Salient Object Detection}
\label{sec:FCN}
We revisit traditional FCN based salient object detection methods on how to extract global-local representations. Normally, a CNN-based encoder (e.g., VGG~\cite{simonyan2014very} and ResNet~\cite{he2016deep}) takes an image as input. Due to the locality of CNN, a stack of convolution layers is applied to the input images. At the same time, the spatial resolution of the feature maps gradually reduces at the end of each stage. Such an operation extends the receptive field of the encoder and reaches a global view in deep layers. This kind of encoder faces three serious problems: 1) features from deep layers have a global view with diluted local details due to too much convolution operations and resolution reduction; 2) features from shallow layers have more local details but lack of semantic understanding to identify salient objects, due to the locality of the convolution layers; 3) features from adjacent layers have limited variances and lead to redundancy when they are densely decoded.

Recent work focuses on strengthening the ability to extract either local details or global semantics. To learn local details, previous works~\cite{zhao2019egnet,wang2019salient,wu2019stacked} focus on features from shallow layers by adding edge supervision. Such low-level features lack a global view and may misguide the whole network. To learn global semantics, attention mechanism is used to enlarge the receptive field. However, previous works~\cite{chen2018reverse,zhang2018progressive} only apply the attention mechanism in deep layers (\eg, the last layer of the encoder). In the following subsections, we demonstrate how to preferably unify local-global features using the transformer.

\subsection{Transformer Encoder}
\label{sec:transformer}
\subsubsection{Image Serialization}
Since the transformer is designed for the 1D sequence in natural language processing tasks, we first map the input 2D images into 1D sequences. Specifically, given an image $x \in \mathcal{R}^{H\times W \times c}$ with height $H$, width $W$ and channel $c$, the 1D sequence $x' \in \mathcal{R}^{L \times C}$ is encoded from the image $x$. A simple idea is to directly reshape the image into a sequence, namely, $L=H\times W$. However, due to the quadratic spatial complexity of self-attention, such an idea leads to enormous GPU memory consumption and computation cost. Inspired by ViT~\cite{dosovitskiy2020image}, we divide the image $x$ into $\frac{HW}{256}$ non-overlapping image patches with a resolution of $16\times 16$. Then we project these patches into tokens with a linear layer, and the sequence length $L$ is $\frac{HW}{256}$. Every token in $x'$ represents a non-overlay $16\times 16$ patch. Such an operation trades off the spatial information and sequence length.

\subsubsection{Transformer}
Taking 1D sequence $x'$ as the input, we use pure transformer as the encoder instead of CNN used in previous works~\cite{Wei_2020_CVPR,Qin_2019_CVPR,Pang_2020_CVPR}. Rather than only having limited receptive field as CNN-based encoder, with the power of modeling global representations in each layer, the transformer usually requires much fewer layers to receive a global receptive. Besides, the transformer encoder has connections across each pair of image patches. Based on these two characteristics, more local detailed information is preserved in our transformer encoder. Specifically, the transformer encoder contains positional encoding and transformer layers with multi-head attention and multi-layer perception.

\textbf{Positional Encoding.} Since the attention mechanism cannot distinguish the positional difference, the first step is to feed the positional information into sequence $x'$ and get the position enhanced feature $F_0$:
\begin{equation}
F_0 = [x^{'}_0, x^{'}_1, ..., x^{'}_L] + E_{pos},
\end{equation}
where $+$ is the addition operation and $E_{pos}$ indicates the positional code which is randomly initialized under truncated Gaussian distribution and is trainable in our method.

\textbf{Transformer Layer.} The transformer encoder contains 17 transformer layers, and each layer contains multi-head self-attention (MSA) and multi-layer perceptron (MLP). The multi-head self-attention is the extension of self-attention (SA):
\begin{equation}
\centering
\begin{split}
Q = F W_q, K = FW_k, V=FW_v,\\
SA(F) = \phi (\frac{QK^T}{\sqrt{d}})V,
\end{split}
\end{equation}
where $F$ indicates the input features of the self-attention, and $W_q, W_k, W_v$ are the weights with the trainable parameters. d is the dimension of $Q, K, V$ and $\phi(\star)$ is the softmax function. To apply multiple attentions in parallel, multi-head self-attention has $m$ independent self-attentions:
\begin{equation}
MSA(F) = SA_1(F)\oplus SA_2(F) \oplus ... \oplus SA_m(F),
\end{equation}
where $\oplus$ indicates the concatenation. To sum up, in the $i_{th}$ transformer layer, the output feature $F_i$ is calculated by:
\begin{equation}
\begin{split}
&\hat{F}_i = MSA(LN(F_{i-1})) + F_{i-1},\\
&F_i = MLP(LN(\hat{F}_i))+\hat{F}_i,
\end{split}
\end{equation}
where $LN(\star)$ indicates layer norm and $F_i$ is the $i_{th}$ layer features of the transformer.

\subsection{Various Decoders}
\begin{figure*}[t]
    \centering
    \includegraphics[width=\linewidth]{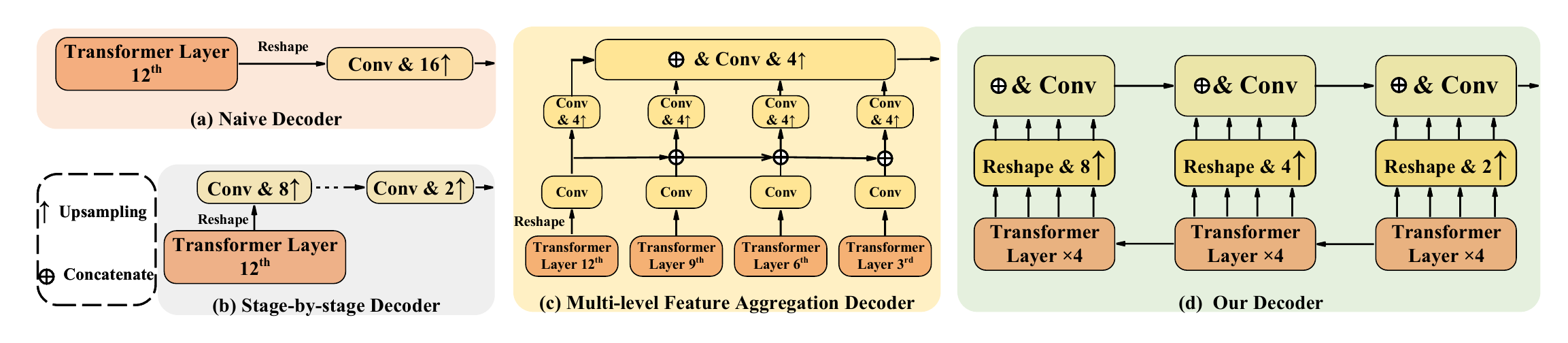}

    \caption{
    	Different types of decoders. (a) Naive Decoder directly upsamples the
output 16$\times$. (b) Stage-by-Stage Decoder upsamples the resolution 2$\times$ in each stage. (c) Multi-level feature Aggregation Decoder sparsely fuses multi-level features. Our decoder (d) densely decodes all transformer features and gradually upsamples to the resolution of inputs. }
    \label{fig:compare}%
\end{figure*}

In this section, we design a global-local decoder to decode the features $F \in R^{\frac{HW}{256}\times C}$ from the transformer encoder and generate the final saliency maps $S^{'} \in R^{H\times W}$ with the same resolution as the inputs. Before exploring the ability of our global-local decoder, we simply introduce the naive decoder, stage-by-stage decoder, and multi-level feature aggregation decoder used in \cite{zheng2020rethinking} (see Fig.~\ref{fig:compare}). To begin with, we first reshape the features $F_i \in R^{\frac{HW}{256}\times C}$ from the transformer into $F^{'}_i \in R^{\frac{H}{16} \times \frac{W}{16} \times C}$.

\textbf{Naive Decoder.} As illustrated in Fig. \ref{fig:compare}(a), a naive idea of decoding transformer features is directly upsampling the outputs of last layer to the same resolution of inputs and generating the saliency maps. Specifically, three conv-batch norm-ReLU layers are applied on $F_{12}$. Then we bilinearly upsample the feature maps 16$\times$, followed by a simple classification layer:
\begin{equation}
\begin{split}
Z = Up(CBR(F^{'}_{12}; \theta_z); 16),\\
S^{'} = \sigma(Conv(Z; \theta_s)),
\end{split}
\end{equation}
where $CBR(\star; \theta)$ indicates convolution-batch norm-ReLU layer with parameters $\theta$. $Up(\star; s)$ is the bilinear operation for upsampling $s\times$ and $\sigma$ is the sigmoid function. $S'$ is the predicted saliency map.

\textbf{Stage-by-Stage Decoder.}
Directly upsampling 16$\times$ at one time like the naive decoder will inject too much noise and lead to coarse spatial details. Inspired by previous FCN-based methods~\cite{Qin_2019_CVPR} in salient object detection, stage-by-stage decoder which upsamples the resolution 2$\times$ in each stage will mitigate the losses of spatial details seen in Fig. \ref{fig:compare}(b). Therefore, there are four stages, and in each stage, a decoder block contains three conv-batch norm-ReLU layers:
\begin{equation}
\begin{split}
&Z_0 = CBR(F^{'}_{12}; \theta_0),\\
&Z_i = CBR(Up(Z_{i-1}; 2); \theta_i),\\
&S^{'} = \sigma(Conv(Z_4; \theta_s)).
\end{split}
\end{equation}

\textbf{Multi-level feature Aggregation.} As illustrated in Fig. \ref{fig:compare}(c), Zheng \etal~\cite{zheng2020rethinking} propose a multi-level feature aggregation to sparsely fuse multi-level features following the similar setting of pyramid network. Rather than having a pyramid shape, the spatial resolutions of features keep the same in the transformer encoder.

Specifically, the decoder take features $F_3, F_6, F_9, F_{12}$ from the $3_{rd}, 6_{th}, 9_{th}, 12_{th}$ layers and upsample them 4$\times$ following several convolutional layers. Then features are top-down aggregated via element-wise addition. Finally, the features from all streams are fused to generate the final saliency maps.

\textbf{Deeply-transformed Decoder.} As illustrated in Fig. \ref{fig:compare}(d), to combine the benefit from the features of each transformer layer and include less noise injection from upsampling, we gradually integrate all transformer layer features and upsample them into the same spatial resolution of the input images. Specifically, we first divide the transformer features into three stages and we upsample them $8\times, 4\times, 2\times$ from the first to the third stage respectively. Note that the upsampling operation includes pixel shuffle ($2^{3-n}\times$) and bilinear upsampling (2$\times$), where $n$ indicates the $n_{th}$ stage. In the $j_{th}$ layer of the $i_{th}$ stage, the salient feature $Z_{i,j}$ comes from the concatenation of the transformer feature $F_{i*4+j}$ from the corresponding layer together with the salient feature from the previous layer. A convolution block is applied after upsampling $F^{'}_{12}$ for $Z_{3,4}$.
\begin{equation}
\small
Z_{i,j} =
\begin{cases}
CBR(Z_{i,j-1}\oplus Up(F^{'}_{i*4+j};2^i); \theta_{i,j}) ~~~~~~~~j=1,2,3,&\\
CBR(Up(Z_{i-1,j};2)\oplus Up(F^{'}_{i*4+j};2^i); \theta_{i,j}) ~~~~j=4.&
\end{cases}
\end{equation}

Then we simply apply a convolutional operation on salient features $Z_{i,j}$ to generate the final saliency maps $S^{'}_{i,j}$. Therefore, in our method, we have twelve output saliency maps:
\begin{equation}
S^{'}_{i,j} = \sigma(Conv(Up(Z_{i,j};2); \theta_{s_{i,j}})).
\end{equation}

\textbf{Loss Function.}
We use standard binary cross-entropy loss as our loss function. Given twelve outputs, we design the final loss as:

\begin{equation}
\begin{split}
l_{bce}(S^{'}, S) &= S \log S^{'} + (1-S)\log(1-S^{'}),\\
L &= \sum\limits_{i=1}^{3} \sum\limits_{j=1}^{4} l_{bce}(S^{'}_{i,j}, S),
\end{split}
\end{equation}
where $\{S^{'}_i\}$ indicate our predicted saliency maps and $S$ indicates the ground truth.

\section{Experiment}
\renewcommand{\arraystretch}{1.5}
\begin{table*}[t]
	\centering
	\resizebox{\textwidth}{!}{
		\begin{threeparttable}
			
			\begin{tabular}{>{\centering}p{1cm}>{\centering}p{0.05cm}|>{\centering}p{0.6cm}>{\centering}p{0.7cm}>{\centering}p{0.7cm}| >{\centering}p{0.6cm}>{\centering}p{0.7cm}>{\centering}p{0.7cm}| >{\centering}p{0.6cm}>{\centering}p{0.7cm}>{\centering}p{0.7cm}| >{\centering}p{0.6cm}>{\centering}p{0.7cm}>{\centering}p{0.7cm}| >{\centering}p{0.6cm}>{\centering}p{0.7cm}>{\centering}p{0.7cm}}
				\toprule
				\multicolumn{2}{c}{\multirow{2}*{Method}}&
				\multicolumn{3}{c}{DUTS-TE}&
				\multicolumn{3}{c}{DUT-OMRON}&
				\multicolumn{3}{c}{HKU-IS}&
				\multicolumn{3}{c}{ECSSD}&
				\multicolumn{3}{c}{PASCAL-S}\cr
				\cmidrule(lr){3-5} \cmidrule(lr){6-8} \cmidrule(lr){9-11} \cmidrule(lr){12-14}\cmidrule(lr){15-17}
				& &MAE~$\downarrow$&$F_{\beta}$~$\uparrow$ &$\mathcal{S}$~$\uparrow$&
				MAE~$\downarrow$ &$F_{\beta}$~$\uparrow$ &$\mathcal{S}$~$\uparrow$&
				MAE~$\downarrow$&$F_{\beta}$~$\uparrow$ &$\mathcal{S}$~$\uparrow$&
				MAE~$\downarrow$&$F_{\beta}$~$\uparrow$ &$\mathcal{S}$~$\uparrow$&
				MAE~$\downarrow$&$F_{\beta}$~$\uparrow$ &$\mathcal{S}$~$\uparrow$ \cr
				\midrule
				DSS &C&  0.056&0.801&0.824& 0.063&0.737&0.790& 0.040&0.889&0.790& 0.052&0.906&0.882& 0.095&0.809&0.797\cr
				\rowcolor{mygray}
				UCF &C& 0.112&0.742&0.782& 0.120&0.698&0.760& 0.062&0.874&0.875& 0.069&0.890&0.883& 0.116&0.791&0.806\cr
				Amulet &C& 0.085&0.751&0.804& 0.098&0.715&0.781& 0.051&0.887&0.886& 0.059&0.905&0.891& 0.100&0.810&0.819\cr
				\rowcolor{mygray}
				BMPM &C& 0.049&0.828&0.862& 0.064&0.734&0.809& 0.039&0.910&0.907& 0.045&0.917&0.911 &0.074&0.836&0.846\cr
				RAS&C&0.059&0.807&0.838&0.062&0.753&0.814&-&-&-&0.056&0.908&0.893&0.104&0.803&0.796\cr
				\rowcolor{mygray}
				PSAM&C&0.041&0.854&0.879&0.057&0.765&0.830&0.034&0.918&0.914&0.040&0.931&0.920&0.075&0.847&0.851\cr
				CPD &C&0.043&0.841&0.869& 0.056&0.754&0.825& 0.034&0.911&0.906& 0.037&0.927&0.918 &0.072&0.837&0.847\cr
				\rowcolor{mygray}
				SCRN &C&0.040&0.864&0.885& 0.056&0.772&0.837& 0.034&0.921&0.916& 0.038&0.937&{0.927}& 0.064&0.858&0.868\cr
				
				BASNet &C&0.048&0.838&0.866& 0.056&0.779&0.836& 0.032&0.919&0.909& 0.037 &0.932&0.916& 0.078&0.836&0.837\cr
				\rowcolor{mygray}
				EGNet &C&0.039&0.866&0.887& 0.053&0.777&0.841& 0.031&0.923&0.918& 0.037 &0.936&0.925& 0.075&0.846&0.853\cr
				MINet &C&0.037&0.865&0.884& 0.056&0.769&0.833& 0.029&0.926&0.919& {0.034}&0.938&0.925& 0.064&0.852&0.856\cr
				\rowcolor{mygray}
				LDF &C&{0.034}&{0.877}&{0.892} &{0.052}&0.782&0.839& {0.028}&0.929&0.920&{ 0.034}&0.938&0.924&
				{0.061}&0.859&0.862\cr
				GateNet &C&0.040&0.869&0.885& 0.055&0.781&0.838& 0.033&0.920&0.915& 0.040&0.933&0.920& 0.069&0.852&0.858 \cr
				\rowcolor{mygray}
				SAC&C&{0.034}&0.882&0.895&{0.052}&{0.804}&{0.849}&{0.026}&{0.935}&{0.925}&{0.031}&{0.945}&{0.931}&0.063&{0.868}&{0.866}\cr
    
				PoolNet+&C &0.039 &0.887 & 0.890&0.056 &0.801 & 0.842 &0.034&0.933  &0.921  &0.040&0.941 &0.925& 0.068&0.869 &0.864\cr
    \rowcolor{mygray}
                RCSBNet&C &0.039 &0.887 & 0.890&0.056 &0.801 & 0.842 &0.034&0.933  &0.921  &0.040&0.941 &0.925& 0.068&0.869 &0.864\cr\hline
    
				Naive &T&0.043&0.855&0.878 &0.059&0.776&0.835&  0.042&0.905&0.903& 0.042&0.927&0.919& 0.068&0.854&0.862\cr
				\rowcolor{mygray}
				SETR &T&0.039&0.869&0.888 &0.056&0.782&0.838& 0.037&0.917&0.912& 0.041&0.930&0.921& {0.062}&{0.867}&0.859\cr
				VST &T&0.037&{0.877}&{0.896}&0.058&{0.800}&{0.850}&0.030&{0.937}&{0.928}&{0.034}&{0.944}&{0.932}&0.067&0.850&{0.873}\cr
\rowcolor{mygray}
                TCFNet&T& 0.031&0.881&0.899& 0.049& 0.802&0.931 &0.027&0.928&0.925 &0.029&0.938&0.930&0.058&0.870&0.866 \cr
				Ours &T&\textbf{\red{0.029}}&\textbf{\red{0.901}}&\textbf{\red{0.912}} &\textbf{\red{0.045}}&\textbf{\red{0.819}}&\textbf{\red{0.865}}&  \textbf{\red{0.026}}&\textbf{\red{0.941}}&\textbf{\red{0.932}}& \textbf{\red{0.028}}&\textbf{\red{0.953}}&\textbf{\red{0.940}}& \textbf{\red{0.054}}&\textbf{\red{0.882}}&\textbf{\red{0.881}}\cr
				\bottomrule
			\end{tabular}
			\end{threeparttable}}
				\caption{Quantitative comparisons with FCN-based (denoted as C) and Transformer-based (denoted as T) salient object detection methods by three evaluation metrics. The best performances are marked in \textbf{\red{Red}}. Simply utilizing a transformer (Naive) achieves comparable performance. Our method significantly outperforms all competitors regardless of what basic component they used (i.e., either FCN or transformer-based).The results of VST$\dag$ come from the paper due to the lack of released code or saliency maps.}
			\label{tab:quantitative}
\end{table*}

\begin{figure*}[t]
	\centering
	\hspace{-5mm}
	\subfloat[Image]{
		\begin{minipage}[b]{0.098\textwidth}
			\centering
			\includegraphics[scale=0.14]{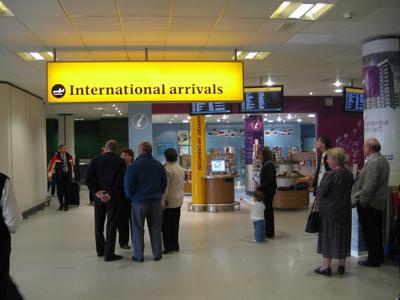}\\
			\includegraphics[scale=0.14]{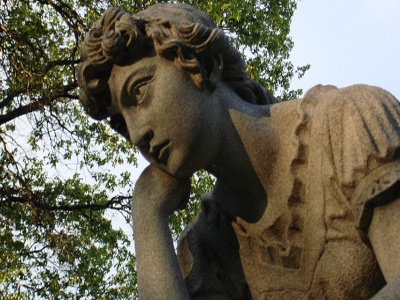} \\
			\includegraphics[scale=0.14]{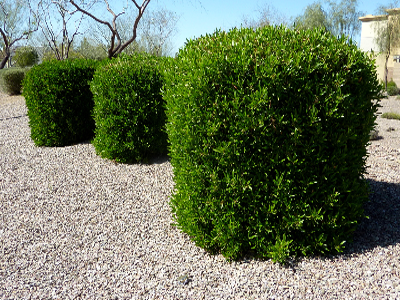} \\
			\includegraphics[scale=0.14]{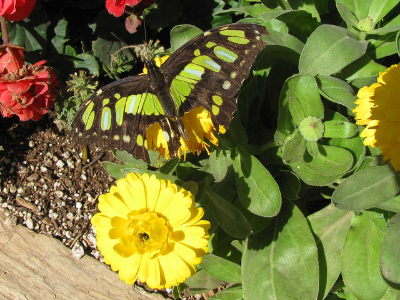} \\
			\includegraphics[scale=0.14]{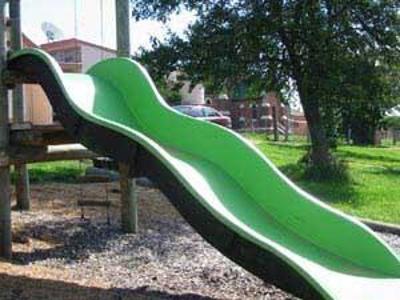}\\
			\includegraphics[scale=0.14]{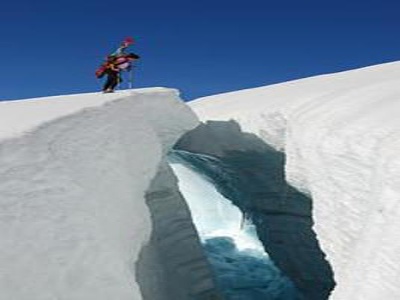} \\
			\includegraphics[scale=0.14]{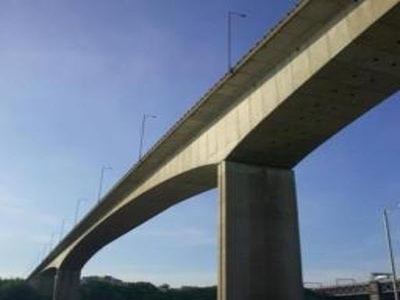} \\
			\includegraphics[scale=0.14]{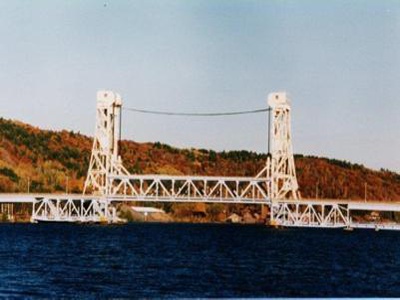}\\
			\includegraphics[scale=0.14]{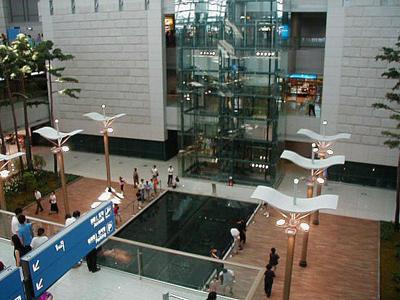}
			
		\end{minipage}
		\label{aaa}
	}
	\subfloat[GT]{
		\begin{minipage}[b]{0.098\textwidth}
			\centering
			\includegraphics[scale=0.14]{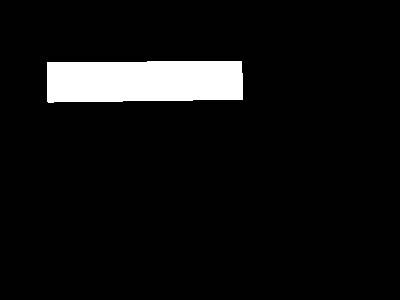}\\
			\includegraphics[scale=0.14]{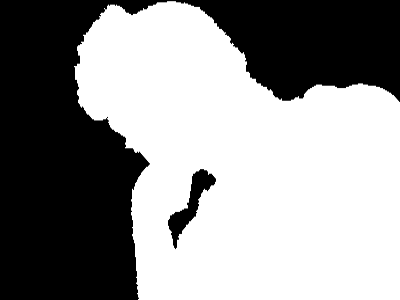} \\
			\includegraphics[scale=0.14]{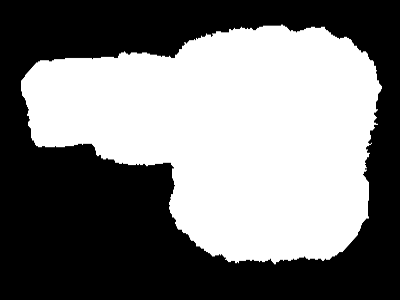}\\
			\includegraphics[scale=0.14]{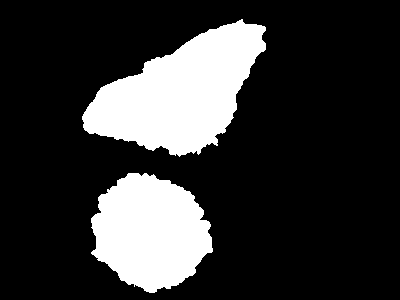} \\
			\includegraphics[scale=0.14]{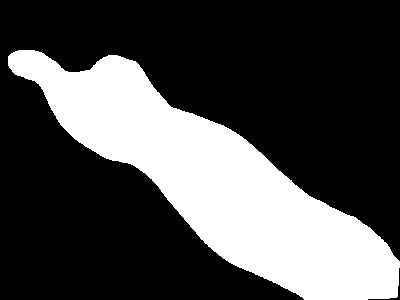}\\
			\includegraphics[scale=0.14]{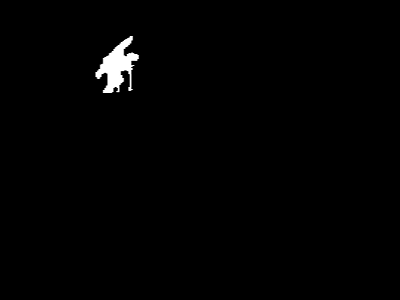} \\
			\includegraphics[scale=0.14]{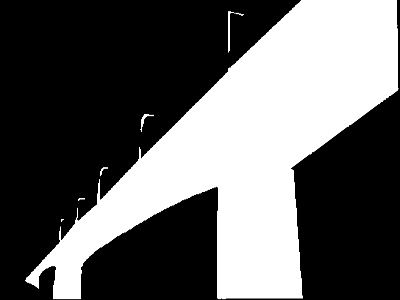}\\
			\includegraphics[scale=0.14]{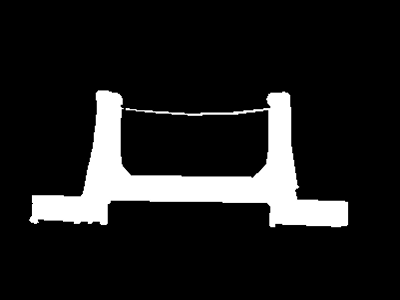} \\
			\includegraphics[scale=0.14]{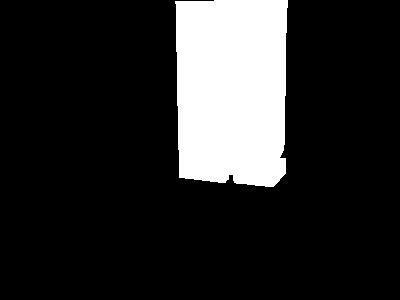}
		\end{minipage}
	}
	\subfloat[Ours~(T)]{
		\begin{minipage}[b]{0.098\textwidth}
			\centering
			\includegraphics[scale=0.14]{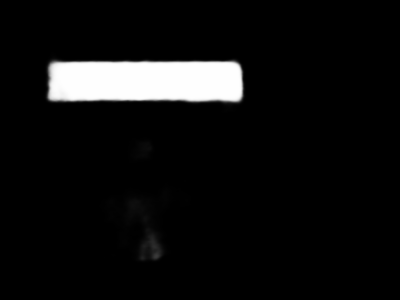}\\
			\includegraphics[scale=0.14]{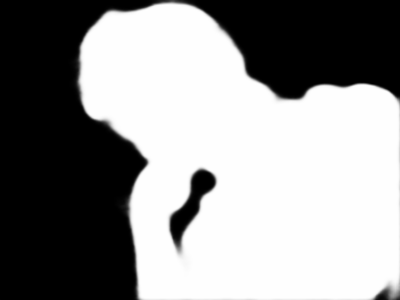}\\
			\includegraphics[scale=0.14]{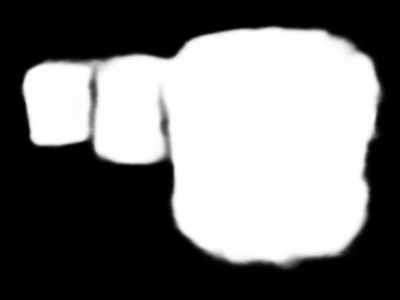}\\
			\includegraphics[scale=0.14]{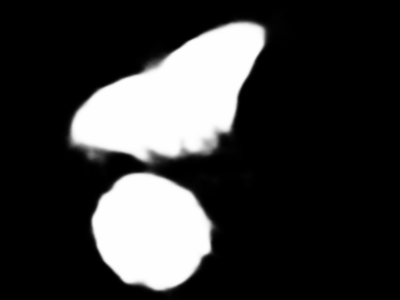}\\
			\includegraphics[scale=0.14]{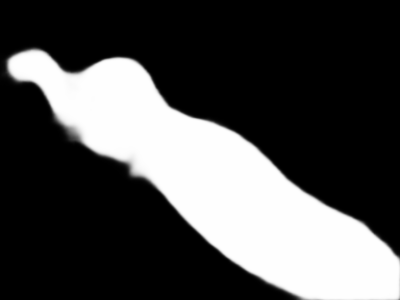}\\
			\includegraphics[scale=0.14]{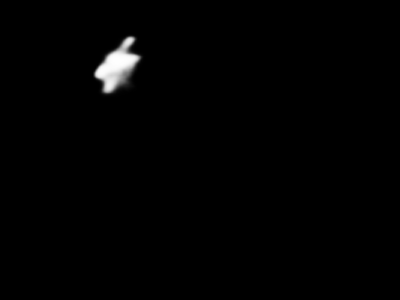}\\
			\includegraphics[scale=0.14]{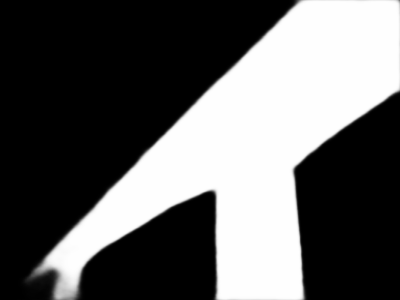}\\
			\includegraphics[scale=0.14]{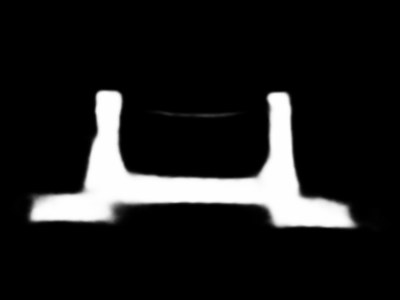}\\
			\includegraphics[scale=0.14]{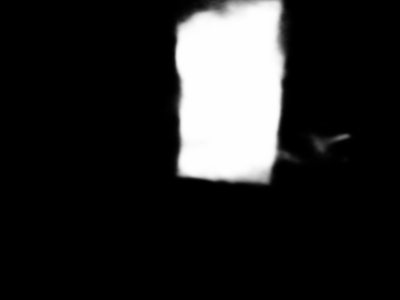}
		\end{minipage}
	}
	\subfloat[SETR~(T)]{
		\begin{minipage}[b]{0.098\textwidth}
			\centering
			\includegraphics[scale=0.14]{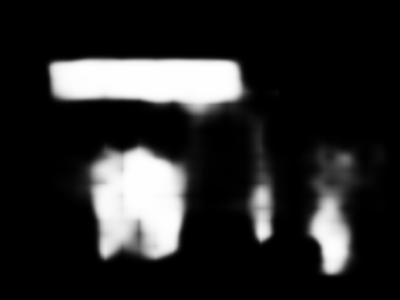}\\
			\includegraphics[scale=0.14]{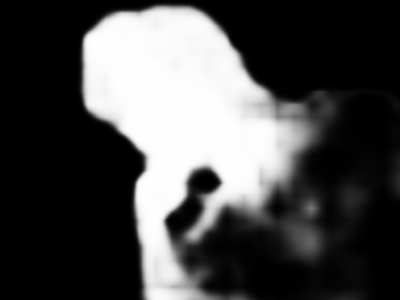} \\
			\includegraphics[scale=0.14]{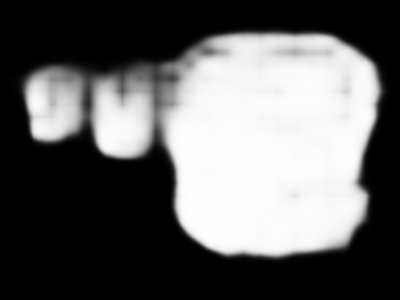} \\
			\includegraphics[scale=0.14]{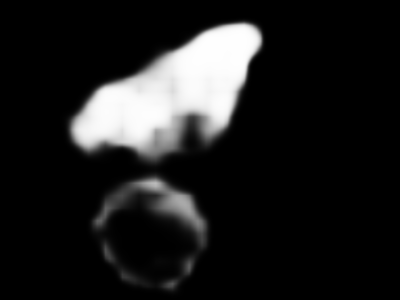}\\
			\includegraphics[scale=0.14]{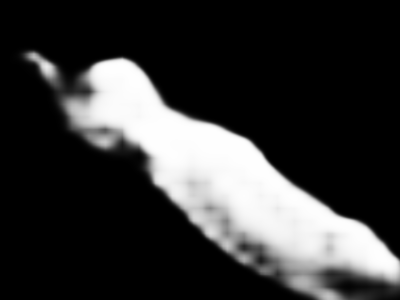}\\
			\includegraphics[scale=0.14]{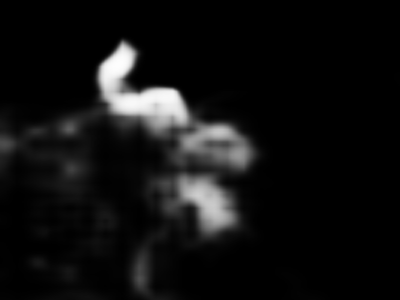}\\
			\includegraphics[scale=0.14]{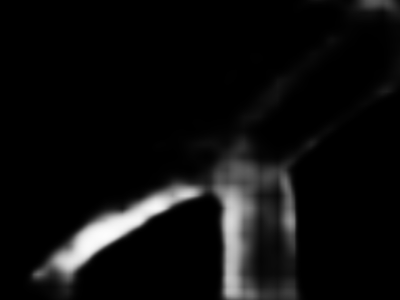}\\
			\includegraphics[scale=0.14]{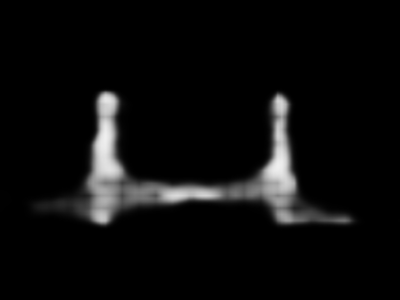}\\
			\includegraphics[scale=0.14]{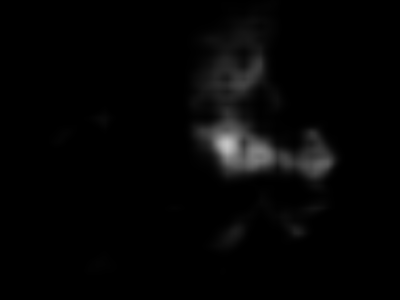}
		\end{minipage}
	}
	\subfloat[GateNet~(C)]{
		\begin{minipage}[b]{0.098\textwidth}
			\centering
			\includegraphics[scale=0.14]{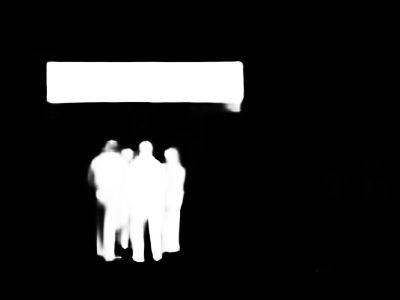} \\
			\includegraphics[scale=0.14]{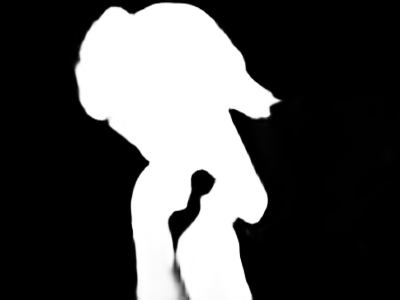} \\
			\includegraphics[scale=0.14]{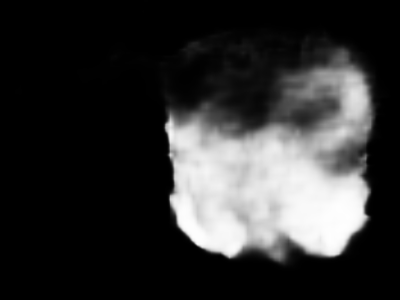} \\
			\includegraphics[scale=0.14]{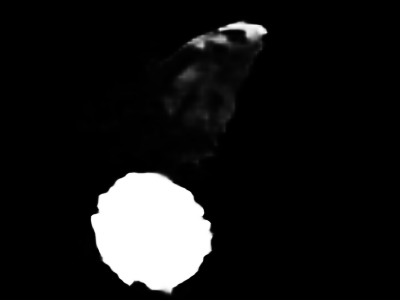} \\
			\includegraphics[scale=0.14]{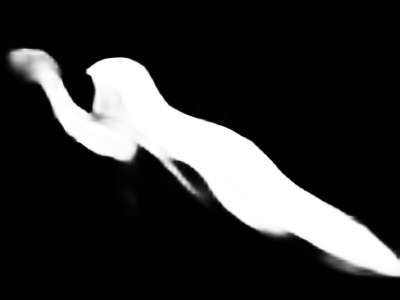} \\
			\includegraphics[scale=0.14]{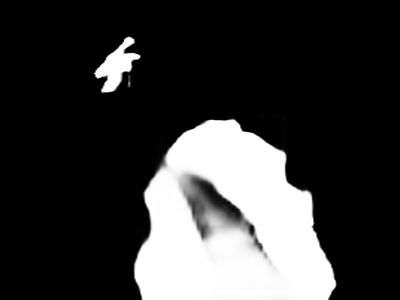} \\
			\includegraphics[scale=0.14]{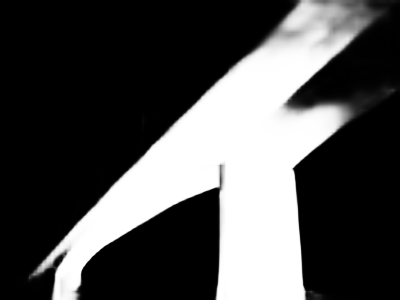} \\
			\includegraphics[scale=0.14]{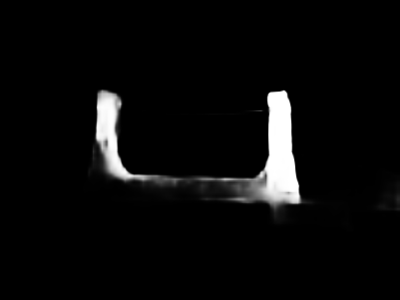} \\
			\includegraphics[scale=0.14]{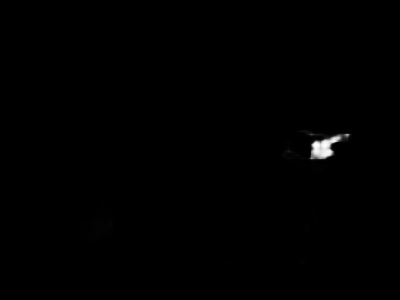}
		\end{minipage}
	}
	\subfloat[LDF~(C)]{
		\begin{minipage}[b]{0.098\textwidth}
			\centering
			\includegraphics[scale=0.14]{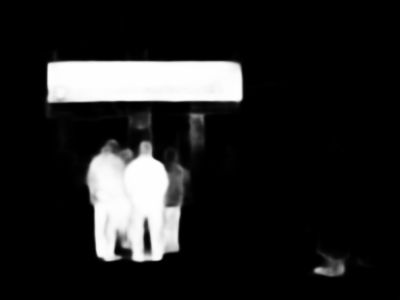} \\
			\includegraphics[scale=0.14]{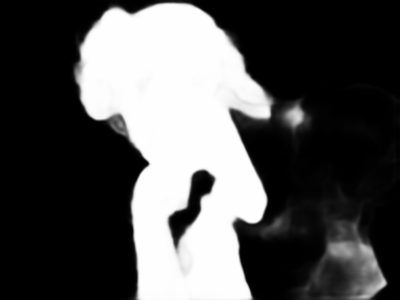}\\
			\includegraphics[scale=0.14]{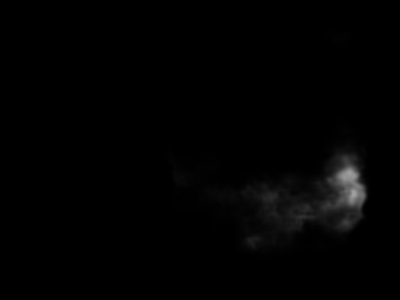}\\
			\includegraphics[scale=0.14]{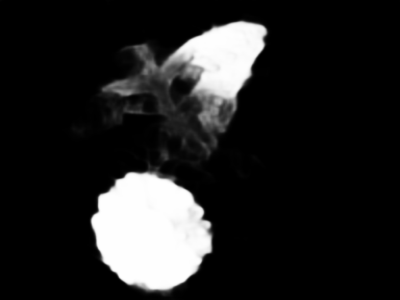} \\
			\includegraphics[scale=0.14]{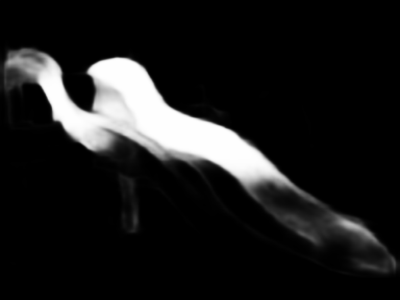}\\
			\includegraphics[scale=0.14]{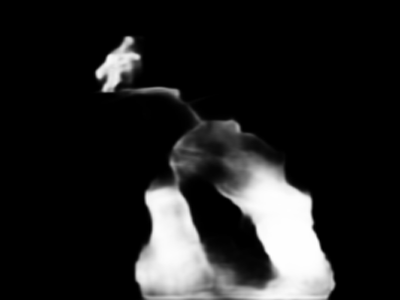}\\
			\includegraphics[scale=0.14]{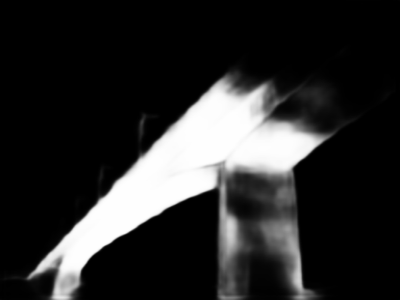} \\
			\includegraphics[scale=0.14]{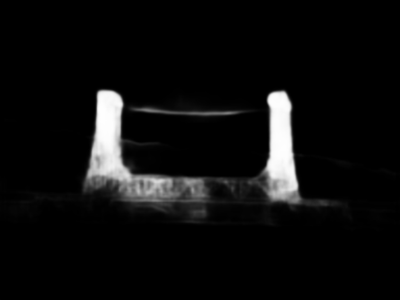}\\
			\includegraphics[scale=0.14]{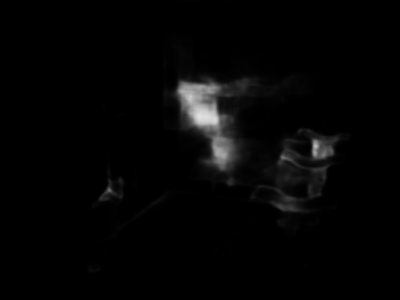}
		\end{minipage}
	}
	\subfloat[EGNet~(C)]{
		\begin{minipage}[b]{0.098\textwidth}
			\centering
			\includegraphics[scale=0.14]{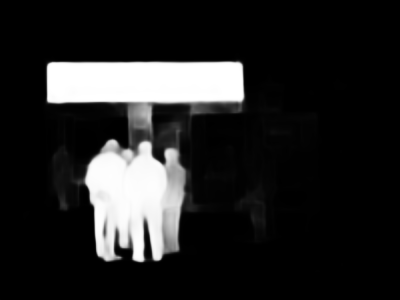} \\
			\includegraphics[scale=0.14]{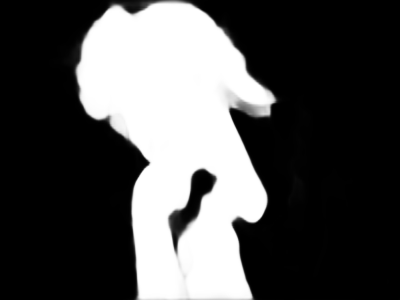} \\
			\includegraphics[scale=0.14]{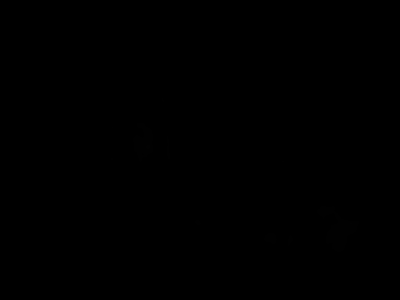} \\
			\includegraphics[scale=0.14]{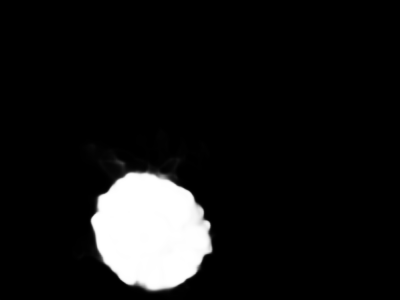} \\
			\includegraphics[scale=0.14]{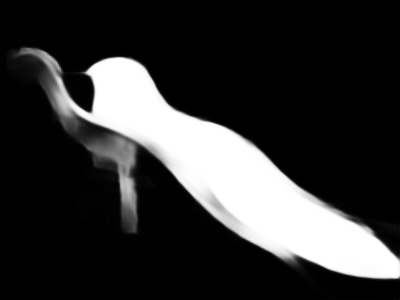} \\
			\includegraphics[scale=0.14]{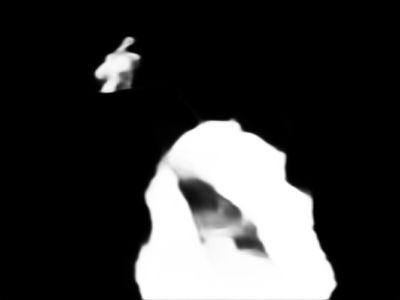} \\
			\includegraphics[scale=0.14]{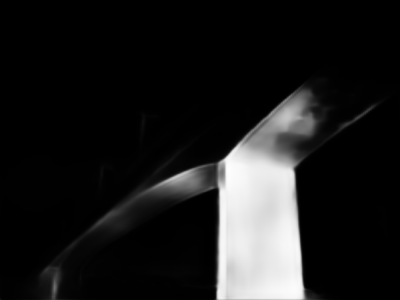} \\
			\includegraphics[scale=0.14]{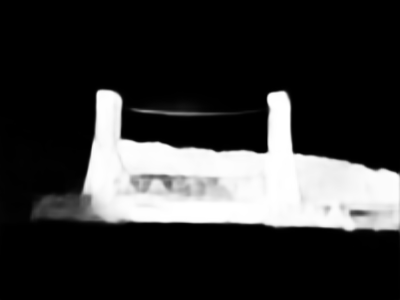} \\
			\includegraphics[scale=0.14]{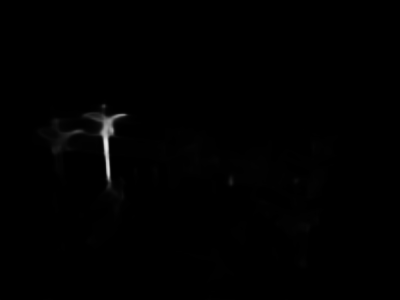}
		\end{minipage}
	}
	\subfloat[BMPM~(C)]{
		\begin{minipage}[b]{0.098\textwidth}
			\centering
			\includegraphics[scale=0.14]{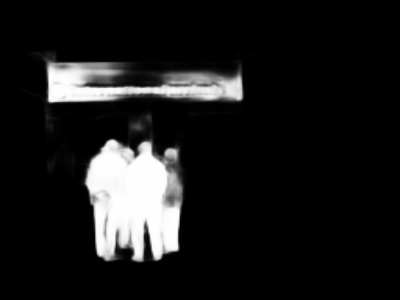} \\
			\includegraphics[scale=0.14]{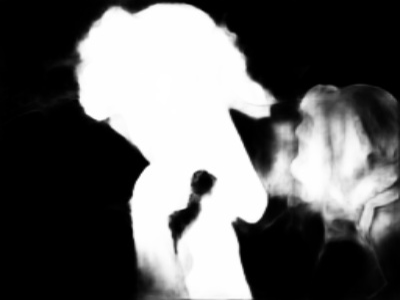} \\
			\includegraphics[scale=0.14]{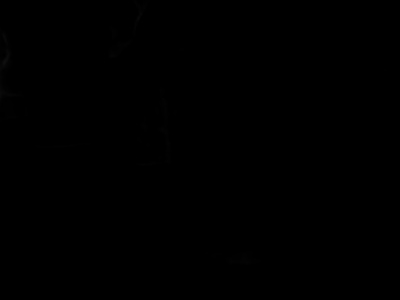} \\
			\includegraphics[scale=0.14]{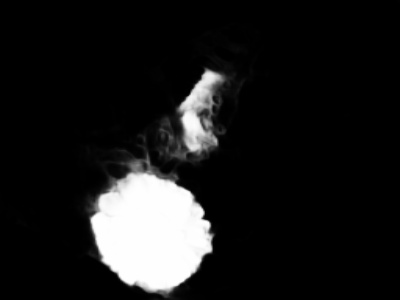} \\
			\includegraphics[scale=0.14]{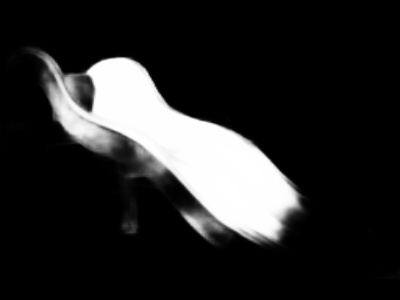} \\
			\includegraphics[scale=0.14]{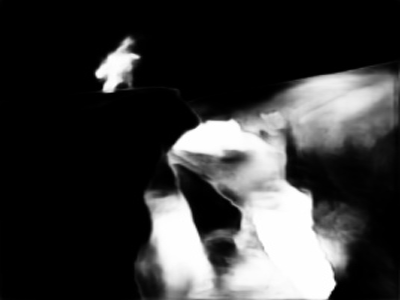} \\
			\includegraphics[scale=0.14]{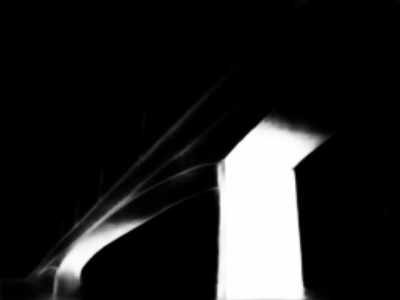} \\
			\includegraphics[scale=0.14]{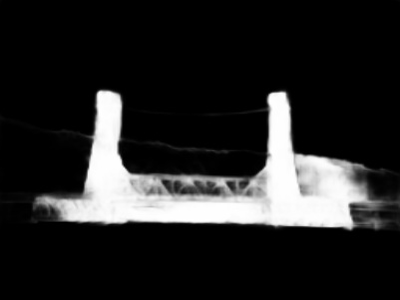} \\
			\includegraphics[scale=0.14]{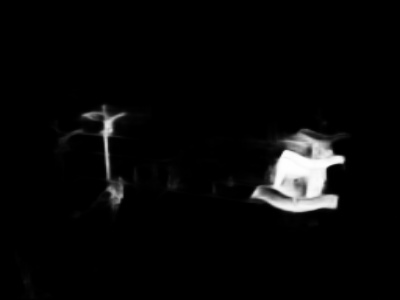}
		\end{minipage}
	}
	\subfloat[DSS~(C)]{
		\begin{minipage}[b]{0.098\textwidth}
			\centering
			\includegraphics[scale=0.14]{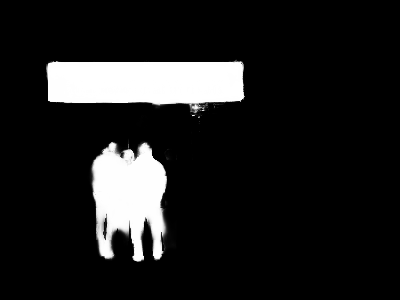} \\
			\includegraphics[scale=0.14]{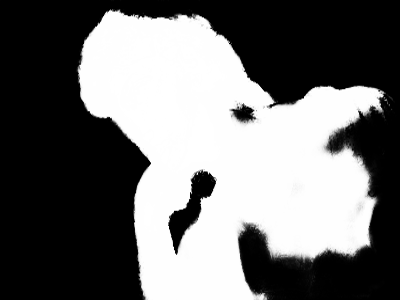} \\
			\includegraphics[scale=0.14]{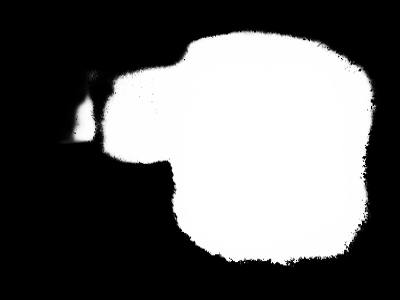} \\
			\includegraphics[scale=0.14]{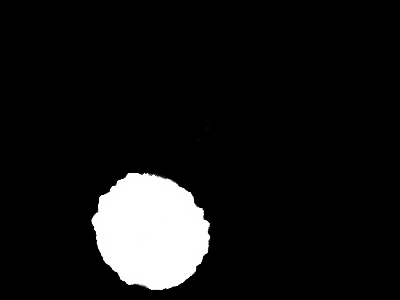} \\
			\includegraphics[scale=0.14]{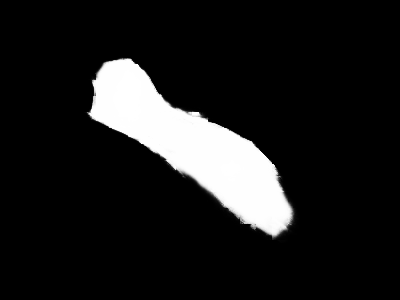} \\
			\includegraphics[scale=0.14]{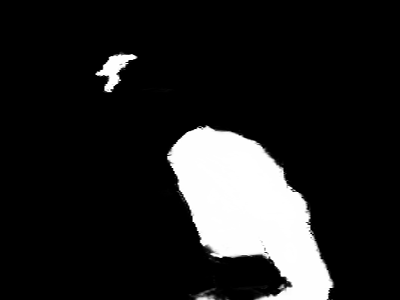} \\
			\includegraphics[scale=0.14]{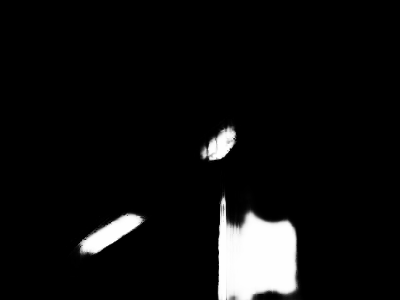} \\
			\includegraphics[scale=0.14]{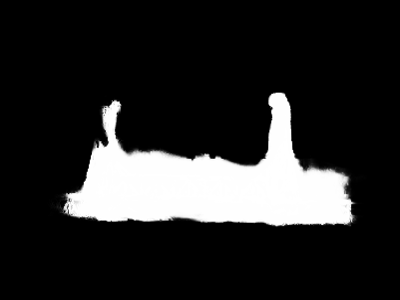} \\
			\includegraphics[scale=0.14]{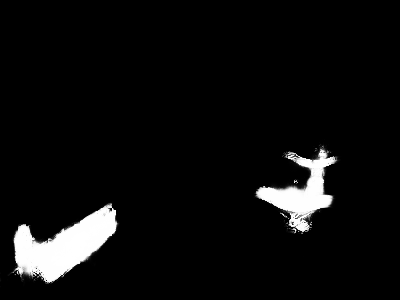}
		\end{minipage}
	}
	\caption{Qualitative comparisons with state-of-the-art methods. Our method provides more visually reasonable saliency maps by accurately locating salient objects and generating sharp boundaries than other transformer-based (denoted as T) and FCN-based methods (denoted as C). }
	\label{Fig:Quantitative}
\end{figure*}

\subsection{Implementation Details}
We train our model on DUTS following the same setting as~\cite{zhao2019egnet}. The transformer encoder is pretrained on ImageNet~\cite{dosovitskiy2020image} and the rest layers are randomly initialized following the default settings in PyTorch. We use SGD with momentum as the optimizer. We set momentum = 0.9, and weight decay = 0.0005. The learning rate starts from 0.001 and gradually decays to 1e-5. We train our model for 40 epochs with a batch size of 8. We adopt vertical and horizontal flipping as the data augmentation techniques. The input images are resized to 384$\times$384. During the inference, we take the output of the last layer as the final prediction.
\subsection{Datasets}
We evaluate our methods on five widely used benchmark datasets: DUT-OMRON~\cite{yang2013saliency}, HKU-IS~\cite{LiYu15}, ECSSD~\cite{shi2015hierarchical}, DUTS~\cite{wang2017}, PASCAL-S~\cite{li2014secrets}. DUT-OMRON contains 5,169 challenging images which usually have complex backgrounds and more than one salient objects. HKU-IS has 4,447 high-quality images with multiple disconnected salient objects in each image. ECSSD has 1,000 meaningful images with various scenes. DUTS is the largest salient object detection dataset including 10,533 training images and 5,019 testing images. PASCAL-S chooses 850 images from PASCAL VOC 2009.

\subsection{Evaluation Metrics}
In this paper, we evaluate our methods under three widely used evaluation metrics: mean absolute error (MAE), weighted F-Measure ($F_{\beta}$), and S-Measure ($S$)~\cite{fan2017structure}.

MAE directly measures the average pixel-wise differences between saliency maps and labels by mean absolute error:

\begin{equation}
MAE(S^{'}, S) = \left| S^{'}-S \right|.
\end{equation}

$F_{\beta}$ evaluates the precision and recall at the same time and use $\beta^2$ to weight precision:
\begin{equation}
F_{\beta} = \frac{(1+\beta^2)\times Precision\times Recall}{\beta^2\times Precision+ Recall},
\end{equation}
where $\beta^2$ is set to 0.3 as in previous work~\cite{zhao2019egnet}.

S-Measure strengthens the structural information of both foreground and background. Specifically, the regional similarity $S_r$ and the object similarity $S_o$ are combined together with weight $\alpha$:
\begin{equation}
S = \alpha*S_o+(1-\alpha)*S_r,
\end{equation}
where $\alpha$ is set to 0.5~\cite{zhao2019egnet}.

\subsection{Comparison with state-of-the-art methods}
We compare our method with 12 state-of-the-art salient object detection methods: DSS~\cite{hou2017deeply}, UCF~\cite{UCF}, Amulet~\cite{Amulet} BMPM~\cite{zhang2018bi},RAS~\cite{chen2018eccv},  PSAM\cite{chang2018pyramid} PoolNet~\cite{liu2019simple}, CPD~\cite{Wu_2019_CVPR}, SCRN~\cite{9010954}, BASNet~\cite{Qin_2019_CVPR}, EGNet~\cite{zhao2019egnet}, MINet~\cite{Pang_2020_CVPR}, LDF~\cite{Wei_2020_CVPR}, GateNet~\cite{zhao2020suppress}, SAC~\cite{hu2020sac}, PoolNet+~\cite{Liu21PamiPoolNet}, RCSBNet~\cite{Ke_2022_WACV}, SETR~\cite{zheng2020rethinking}, VST~\cite{DBLP:journals/corr/abs-2104-12099}, TCFNet~\cite{YAO2023342}, and Naive (Transformer encoder with the naive decoder). The saliency maps are provided by their authors or calculated by the released code except SETR and Naive which are implemented by ourselves. Besides, all results are evaluated with the same evaluation code.

\textbf{Quantitative Comparison.} We report the quantitative results in Tab. \ref{tab:quantitative}. As can be seen from these results, with out any post-processing, our method outperforms all compared methods using FCN-based or transformer-based architecture by a large margin on all evaluation metrics. In particular, the performance is averagely improved 12.17\% over the second best method LDF in terms of MAE. The superior performance on both easy and difficult datasets shows that our method does well on both simple and complex scenes. Another interesting finding is that transformer is powerful to generate rather good saliency maps with ``Naive'' decoder, although it cannot outperform state-of-the-art FCN-based method (i.e., SAC). Further investigating the potential for transformer may keep enhancing the performance on salient object detection task.

\textbf{Qualitative Comparison.} The qualitative results are shown in Fig. \ref{Fig:Quantitative}. With the power of modeling long range dependency over the whole image, transformer-based methods are able to capture the salient object more accurately. For example, in the sixth row of Fig. \ref{Fig:Quantitative}, our method can accurately capture the salient object without being deceived by the ice hill while FCN-based methods all fail in this case. Besides, saliency maps predicted by our method are more complete, and aligned with the ground truths.

\subsection{Ablation Study}		
In this subsection, we mainly evaluate the effectiveness of dense connections and different upsampling strategies in our decoder design on two challenging datasets: DUTS-TE and DUT-OMRON.

\textbf{Effectiveness of Dense Connections.}
\begin{figure}[t]
	\centering
	\setlength{\tabcolsep}{.5pt}
	\renewcommand{\arraystretch}{.5}
	\begin{tabular}{ccccccc}
		\includegraphics[width=.138\linewidth]{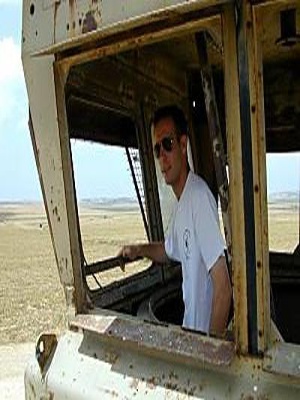}&
		\includegraphics[width=.138\linewidth]{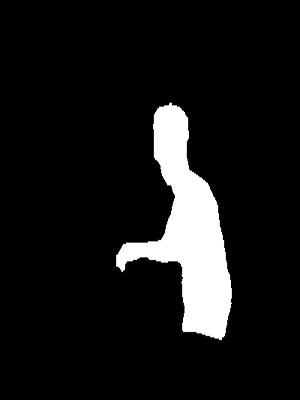}&
		\includegraphics[width=.138\linewidth]{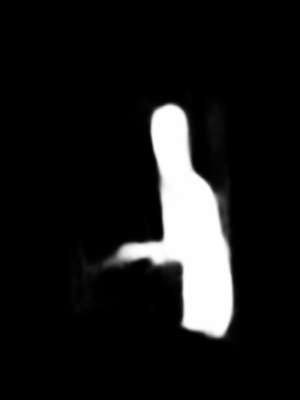}&
		\includegraphics[width=.138\linewidth]{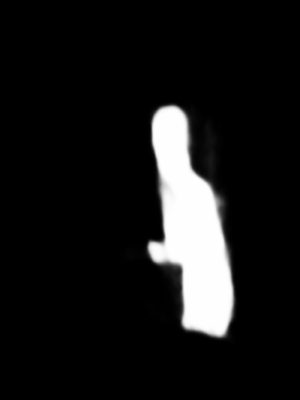}&
		\includegraphics[width=.138\linewidth]{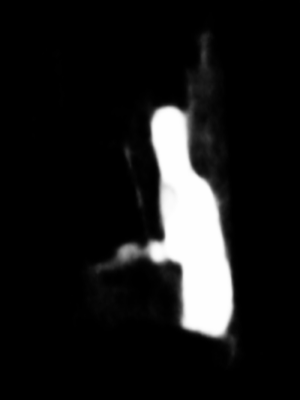}&
		\includegraphics[width=.138\linewidth]{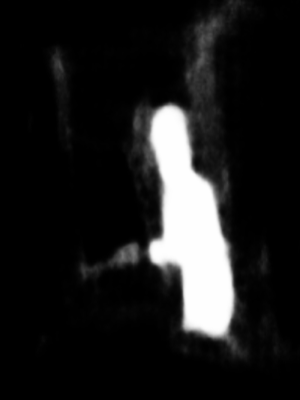}&
		\includegraphics[width=.138\linewidth]{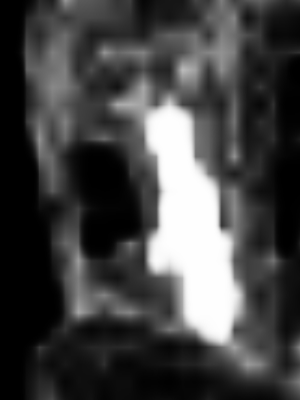}\\
		
		\includegraphics[width=.138\linewidth]{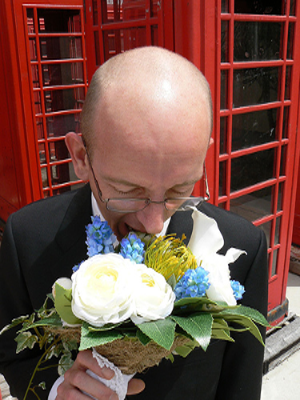}&
		\includegraphics[width=.138\linewidth]{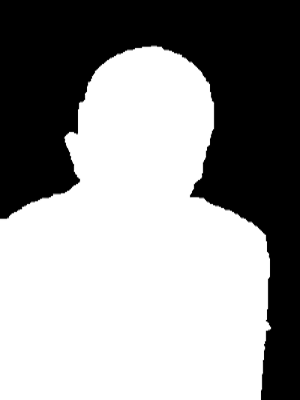}&
		\includegraphics[width=.138\linewidth]{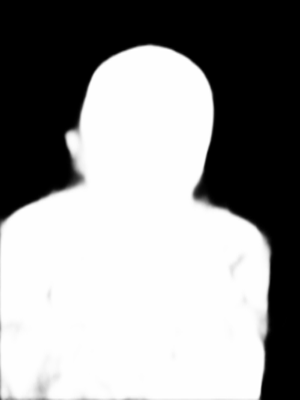}&
		\includegraphics[width=.138\linewidth]{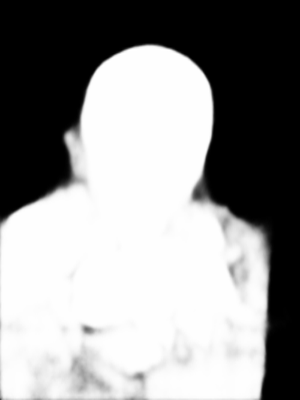}&
		\includegraphics[width=.138\linewidth]{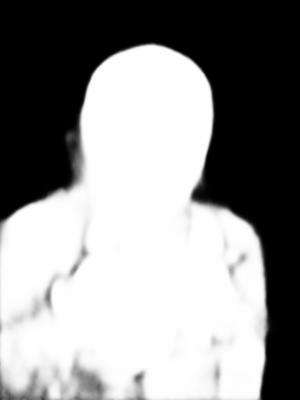}&
		\includegraphics[width=.138\linewidth]{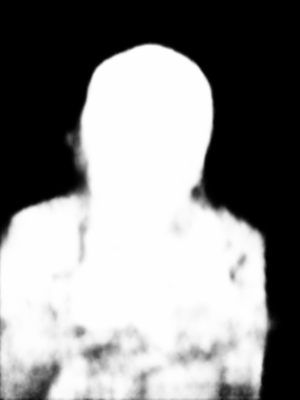}&
		\includegraphics[width=.138\linewidth]{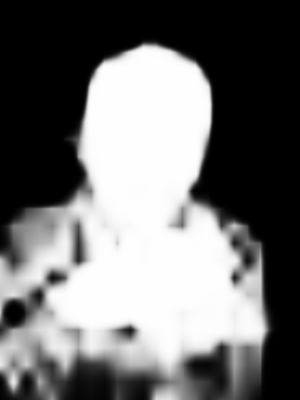}\\

		\includegraphics[width=.138\linewidth]{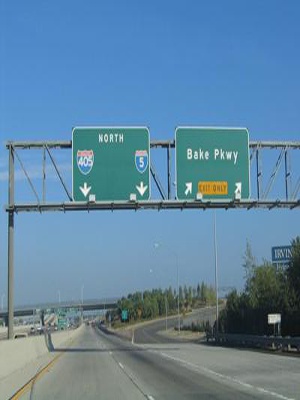}&
		\includegraphics[width=.138\linewidth]{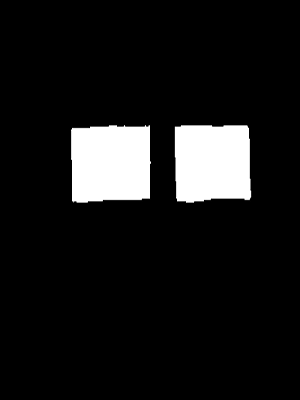}&
		\includegraphics[width=.138\linewidth]{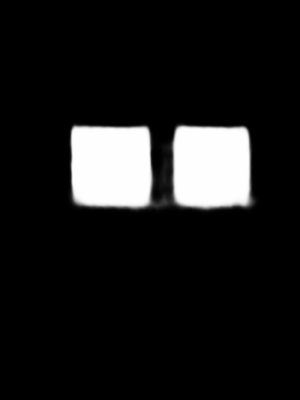}&
		\includegraphics[width=.138\linewidth]{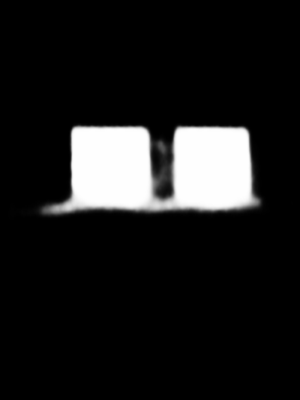}&
		\includegraphics[width=.138\linewidth]{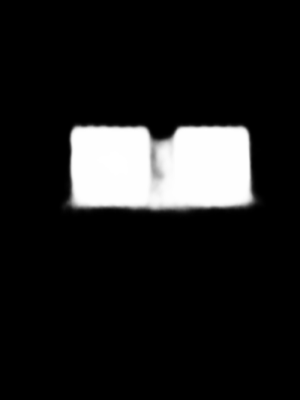}&
		\includegraphics[width=.138\linewidth]{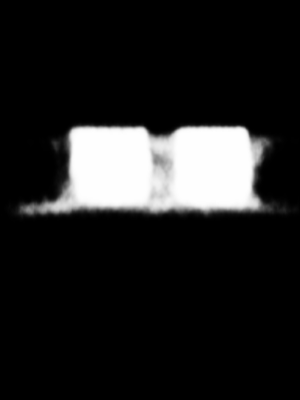}&
		\includegraphics[width=.138\linewidth]{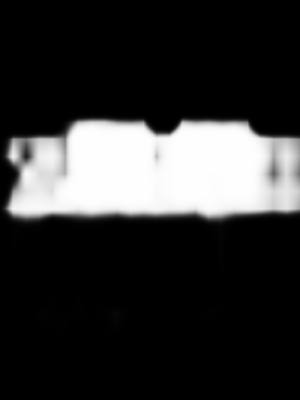}\\

		Input & GT & Dens.4 & Dens.3 & Dens.2 &Dens.1 & Dens.0\\
		&&Ours& & & &Naive
	\end{tabular}
	\caption{The effect of dense connection during decoding. We gradually increase the density of connection between transformer output features and decoding convolution layer in all stages. The naive decoder (i.e., density 0) predicts saliency maps with a lot of noises, which is severely influenced by the background. As the density increasing, the saliency maps become more accurate and sharp, especially near the boundaries.}

	\label{fig:density}
\end{figure}
\renewcommand{\arraystretch}{1.2}
\begin{table}[t]
\centering
\resizebox{0.48\textwidth}{!}{
	\begin{threeparttable}
		\fontsize{8}{8}\selectfont
		\begin{tabular}{c|cccccc}
			\toprule
			Density &\multicolumn{3}{c}{ DUTS-TE}&\multicolumn{3}{c}{ DUT-OMRON}\cr
			\cmidrule(lr){2-4} \cmidrule(lr){5-7}
		 &MAE~$\downarrow$ &$F_{\beta}$ ~$\uparrow$&S~$\uparrow$ &MAE~$\downarrow$ &$F_{\beta}$~$\uparrow$ &S~$\uparrow$  \cr
			\midrule
			0&0.043 &0.855&0.878 &0.059&0.776&0.835 \cr
			\rowcolor{mygray}
			1&0.033 &0.889&0.902 &0.052&0.802&0.852 \cr
			2&0.031 &0.895&0.907 &0.050&0.807&0.857 \cr
			\rowcolor{mygray}
			3&0.030 &0.899&0.910 &0.047&0.813&0.862 \cr
			4&\red{0.029}&\red{0.901}&\red{0.912} &\red{0.045}&\red{0.819}&\red{0.865} \cr
			\bottomrule
		\end{tabular}
	\end{threeparttable}}
\caption{Quantitative evaluation on dense connection during decoding. As can be seen, adding connections can bring significant performance gain compared to the naive one (i.e., density=0). As the density increasing, the performance gradually improves on all three evaluation metrics. The best performances are marked as \red{red}. }

\label{tab:density}
\end{table}
We first conduct an experiment to examine the effect of densely decoding the features from each transformer encoding layer. Starting from simply decoding the features of last transformer (i.e., naive decoder), we gradually add the fuse connections in each stage to increase the density until reaching our final decoder, denoted as density 0, 1, 2, 3. For example, the density 3 means that there are three connections (i.e., from $Z_{i,4}, Z_{i,3}, Z_{i,2}$) added in each stage and only one (i.e., $Z_{i,4}$) for density 1.

The qualitative results are illustrated in Fig. \ref{fig:density}, and quantitative results are shown in Tab.~\ref{tab:density}. When we use the naive decoder without any connection, the results are much worse with lots of noises. The saliency maps are more sharp and accurate with the increase of decoding density, and the performances also gradually increase on three evaluation metrics.
As each layer captures a global view of its previous layer, the representation differences are maximized among different transformer features. Thus, each transformer feature can contribute individually to the predicted saliency maps.

\textbf{Effectiveness of Upsampling Strategy.}
\renewcommand{\arraystretch}{1.5}
\begin{table}[t]
	\centering
	\resizebox{0.48\textwidth}{!}{
		\begin{threeparttable}
			\fontsize{8}{8}\selectfont
			\begin{tabular}{c|cccccc}
				\toprule
				Upsample &\multicolumn{3}{c}{ DUTS-TE}&\multicolumn{3}{c}{ DUT-OMRON}\cr
				\cmidrule(lr){2-4} \cmidrule(lr){5-7}
				Strategy&MAE~$\downarrow$ &$F_{\beta}$ ~$\uparrow$&S~$\uparrow$ &MAE~$\downarrow$ &$F_{\beta}$~$\uparrow$ &S~$\uparrow$  \cr
				\midrule
				\multicolumn{7}{c}{Density set to 0}\cr
				\midrule
				16$\times$&0.043&0.855&0.878 &0.059&0.776&0.835 \cr
				Four 2$\times$&0.041 &0.861&0.883& 0.059&0.775&0.836 \cr
				\midrule
				\multicolumn{7}{c}{Density set to 1}\cr
				\midrule
				16$\times$&0.036 &0.875&0.896& 0.052&0.794&0.850 \cr
				4$\times$ and 4$\times$&0.034 &0.886&0.900& 0.053&0.801&0.850 \cr
				Ours&0.033 &0.889&0.902 &0.052&0.802&0.852 \cr
				\midrule
				\multicolumn{7}{c}{Density set to 4}\cr
				\midrule
				16$\times$&0.034 &0.882&0.902&0.051&0.799&0.846 \cr
				4$\times$ and 4$\times$&0.030 &0.900&0.911 &0.047&0.814&0.862 \cr
				Ours&0.029&0.901&0.912 &0.045&0.819&0.865 \cr
				\bottomrule
			\end{tabular}
	\end{threeparttable}}
	\caption{Qualitative evaluations on the upsampling strategies under different density settings. Gradually upsampling leads to better performance than the naive one and our upsampling strategy outperforms others under different density settings.}
	\label{tab:upsample}
\end{table}
To evaluate the effect of our gradually upsampling strategy used in the decoder, we compare different upsampling strategies on different density settings.
1) The Naive decoder (denoted as 16$\times$) simply upsamples the last transformer features 16$\times$ for prediction. 2) When density is set to 0, we test an upsampling strategy in stage-by-stage decoder (denoted as Four 2$\times$), where the output features of each stage are upsampled 2$\times$ before sending to the next stage and there are four $2\times$ upsampling operations in total. 3) A sampling strategy (denoted as 4$\times$ and 4$\times$) used in SETR~\cite{zheng2020rethinking} upsamples each transformer features 4$\times$, and another 4$\times$ is applied to obtain the final predicted saliency map. 4) Our decoder (\textbf{Ours}) upsamples transformer features 8$\times$, 4$\times$, 2$\times$ for the layers connected to stage 1, stage 2, stage 3 respectively.

The quantitative results are reported in Tab \ref{tab:upsample} with some interesting findings. Gradually upsampling always performs better than the naive one with a single $16\times$ upsampling operation. With the help of our upsampling strategy, the dense decoding manifests its advanced performance, with less noise injected during decoding. Directly decoding the output feature from last transformer in Stage mode only leads to limited performance, further indicating the importance of our dense connection.

\subsection{The Statistics of Timing}
Although the computational cost of transformer encoder may be high, thanks to the parallelism of attention architecture, the inference speed is still comparable to recent methods, shown in Tab.~\ref{tab:running time}. Besides, as a pioneer work, we want to highlight the power of transformer on salient object detection (SOD) task and leave time complexity reduction as the future work.
\begin{table}[H]
	\captionsetup{font=small}
	\centering
	\small
	\resizebox{0.48\textwidth}{!}{
		\begin{tabular}{ccccccc}
			\hline
			Method&DSS&R3Net &BASNet &SCRN& EGNet &Ours  \\
			\hline
			Speed(FPS)&24&19&28&16&7&15 \\
			\hline
	\end{tabular}}
	\caption{Running time of different methods. We test all methods on a single RTX2080Ti.}
	\label{tab:running time}
\end{table}

\section{Conclusion}
In this paper, we explore the unified learning of global-local representations in salient object detection with an attention-based model using transformers. With the power of modeling long-range dependency in transformer, we overcome the limitations caused by the locality of CNN in previous salient object detection methods. We propose an effective decoder to densely decode transformer features and gradually upsample to predict saliency maps. It fully uses all transformer features due to the high representation differences and less redundancy among different transformer features. We adopt a gradual upsampling strategy to keep less noise injection during decoding to magnify the ability to locate accurate salient regions. We conduct experiments on five widely used datasets and three evaluation metrics to demonstrate that our method significantly outperforms state-of-the-art methods by a large margin.

\bibliographystyle{IEEEtran}
\bibliography{ref}

\begin{thebibliography}{10}
\providecommand{\url}[1]{#1}
\csname url@samestyle\endcsname
\providecommand{\newblock}{\relax}
\providecommand{\bibinfo}[2]{#2}
\providecommand{\BIBentrySTDinterwordspacing}{\spaceskip=0pt\relax}
\providecommand{\BIBentryALTinterwordstretchfactor}{4}
\providecommand{\BIBentryALTinterwordspacing}{\spaceskip=\fontdimen2\font plus
\BIBentryALTinterwordstretchfactor\fontdimen3\font minus \fontdimen4\font\relax}
\providecommand{\BIBforeignlanguage}[2]{{%
\expandafter\ifx\csname l@#1\endcsname\relax
\typeout{** WARNING: IEEEtran.bst: No hyphenation pattern has been}%
\typeout{** loaded for the language `#1'. Using the pattern for}%
\typeout{** the default language instead.}%
\else
\language=\csname l@#1\endcsname
\fi
#2}}
\providecommand{\BIBdecl}{\relax}
\BIBdecl

\bibitem{zhao2020suppress}
X.~Zhao, Y.~Pang, L.~Zhang, H.~Lu, and L.~Zhang, ``Suppress and balance: A simple gated network for salient object detection,'' in \emph{ECCV}, 2020, pp. 35--51.

\bibitem{mechrez2019saliency}
R.~Mechrez, E.~Shechtman, and L.~Zelnik-Manor, ``Saliency driven image manipulation,'' \emph{Machine Vision and Applications}, vol.~30, no.~2, pp. 189--202, 2019.

\bibitem{luo2021weakly}
W.~Luo, M.~Yang, and W.~Zheng, ``Weakly-supervised semantic segmentation with saliency and incremental supervision updating,'' \emph{Pattern Recognition}, p. 107858, 2021.

\bibitem{wang2022looking}
W.~Wang, G.~Sun, and L.~Van~Gool, ``Looking beyond single images for weakly supervised semantic segmentation learning,'' \emph{IEEE Transactions on Pattern Analysis and Machine Intelligence}, 2022.

\bibitem{wang2017saliency}
W.~Wang, J.~Shen, R.~Yang, and F.~Porikli, ``Saliency-aware video object segmentation,'' \emph{IEEE transactions on pattern analysis and machine intelligence}, vol.~40, no.~1, pp. 20--33, 2017.

\bibitem{craye2016environment}
C.~Craye, D.~Filliat, and J.-F. Goudou, ``Environment exploration for object-based visual saliency learning,'' in \emph{ICRA}.\hskip 1em plus 0.5em minus 0.4em\relax IEEE, 2016, pp. 2303--2309.

\bibitem{yang2014salient}
Y.~Yang, J.~Yang, J.~Yan, S.~Liao, D.~Yi, and S.~Z. Li, ``Salient color names for person re-identification,'' in \emph{ECCV}.\hskip 1em plus 0.5em minus 0.4em\relax Springer, 2014, pp. 536--551.

\bibitem{zhao2016person}
R.~Zhao, W.~Oyang, and X.~Wang, ``Person re-identification by saliency learning,'' \emph{IEEE TPAMI}, vol.~39, no.~2, pp. 356--370, 2016.

\bibitem{wang2018deep}
W.~Wang, J.~Shen, and H.~Ling, ``A deep network solution for attention and aesthetics aware photo cropping,'' \emph{IEEE transactions on pattern analysis and machine intelligence}, vol.~41, no.~7, pp. 1531--1544, 2018.

\bibitem{cheng2014global}
M.-M. Cheng, N.~J. Mitra, X.~Huang, P.~H. Torr, and S.-M. Hu, ``Global contrast based salient region detection,'' \emph{IEEE TPAMI}, vol.~37, no.~3, pp. 569--582, 2014.

\bibitem{r1_1}
C.~Gong, D.~Tao, W.~Liu, S.~J. Maybank, M.~Fang, K.~Fu, and J.~Yang, ``Saliency propagation from simple to difficult,'' in \emph{Proceedings of the IEEE conference on computer vision and pattern recognition}, 2015, pp. 2531--2539.

\bibitem{einhauser2003does}
W.~Einh{\"a}user and P.~K{\"o}nig, ``Does luminance-contrast contribute to a saliency map for overt visual attention?'' \emph{European Journal of Neuroscience}, vol.~17, no.~5, pp. 1089--1097, 2003.

\bibitem{he2014saliency}
S.~He and R.~W. Lau, ``Saliency detection with flash and no-flash image pairs,'' in \emph{ECCV}.\hskip 1em plus 0.5em minus 0.4em\relax Springer, 2014, pp. 110--124.

\bibitem{r1_4}
K.~Fu, C.~Gong, I.~Y.-H. Gu, and J.~Yang, ``Normalized cut-based saliency detection by adaptive multi-level region merging,'' \emph{IEEE Transactions on Image Processing}, vol.~24, no.~12, pp. 5671--5683, 2015.

\bibitem{long2015fully}
J.~Long, E.~Shelhamer, and T.~Darrell, ``Fully convolutional networks for semantic segmentation,'' in \emph{CVPR}, 2015, pp. 3431--3440.

\bibitem{li2016deep}
G.~Li and Y.~Yu, ``Deep contrast learning for salient object detection,'' in \emph{CVPR}, 2016, pp. 478--487.

\bibitem{wang2017stagewise}
T.~Wang, A.~Borji, L.~Zhang, P.~Zhang, and H.~Lu, ``A stagewise refinement model for detecting salient objects in images,'' in \emph{ICCV}, 2017, pp. 4019--4028.

\bibitem{wang2016saliency}
L.~Wang, L.~Wang, H.~Lu, P.~Zhang, and X.~Ruan, ``Saliency detection with recurrent fully convolutional networks,'' in \emph{ECCV}.\hskip 1em plus 0.5em minus 0.4em\relax Springer, 2016, pp. 825--841.

\bibitem{zhang2018bi}
L.~Zhang, J.~Dai, H.~Lu, Y.~He, and G.~Wang, ``A bi-directional message passing model for salient object detection,'' in \emph{CVPR}, 2018, pp. 1741--1750.

\bibitem{vaswani2017attention}
A.~Vaswani, N.~Shazeer, N.~Parmar, J.~Uszkoreit, L.~Jones, A.~N. Gomez, L.~Kaiser, and I.~Polosukhin, ``Attention is all you need,'' \emph{arXiv preprint arXiv:1706.03762}, 2017.

\bibitem{dosovitskiy2020image}
A.~Dosovitskiy, L.~Beyer, A.~Kolesnikov, D.~Weissenborn, X.~Zhai, T.~Unterthiner, M.~Dehghani, M.~Minderer, G.~Heigold, S.~Gelly \emph{et~al.}, ``An image is worth 16x16 words: Transformers for image recognition at scale,'' \emph{arXiv preprint arXiv:2010.11929}, 2020.

\bibitem{reynolds2003interacting}
J.~H. Reynolds and R.~Desimone, ``Interacting roles of attention and visual salience in v4,'' \emph{Neuron}, vol.~37, no.~5, pp. 853--863, 2003.

\bibitem{morgan1943physiological}
C.~T. Morgan, ``Physiological psychology.'' 1943.

\bibitem{desimone1995neural}
R.~Desimone and J.~Duncan, ``Neural mechanisms of selective visual attention,'' \emph{Annual review of neuroscience}, vol.~18, no.~1, pp. 193--222, 1995.

\bibitem{liu2018picanet}
N.~Liu, J.~Han, and M.-H. Yang, ``Picanet: Learning pixel-wise contextual attention for saliency detection,'' in \emph{CVPR}, 2018, pp. 3089--3098.

\bibitem{10013775}
S.~Ren, W.~Liu, J.~Jiao, G.~Han, and S.~He, ``Edge distraction-aware salient object detection,'' \emph{IEEE MultiMedia}, vol.~30, no.~3, pp. 63--73, 2023.

\bibitem{zhang2018progressive}
X.~Zhang, T.~Wang, J.~Qi, H.~Lu, and G.~Wang, ``Progressive attention guided recurrent network for salient object detection,'' in \emph{CVPR}, 2018, pp. 714--722.

\bibitem{he2017delving}
S.~He, J.~Jiao, X.~Zhang, G.~Han, and R.~W. Lau, ``Delving into salient object subitizing and detection,'' in \emph{ICCV}, 2017, pp. 1059--1067.

\bibitem{ren2020tenet}
S.~Ren, C.~Han, X.~Yang, G.~Han, and S.~He, ``Tenet: Triple excitation network for video salient object detection,'' in \emph{ECCV}.\hskip 1em plus 0.5em minus 0.4em\relax Springer, 2020, pp. 212--228.

\bibitem{Ren_2021_CVPR}
S.~Ren, W.~Liu, Y.~Liu, H.~Chen, G.~Han, and S.~He, ``Reciprocal transformations for unsupervised video object segmentation,'' in \emph{CVPR}, June 2021, pp. 15\,455--15\,464.

\bibitem{wang2020learning}
B.~Wang, W.~Liu, G.~Han, and S.~He, ``Learning long-term structural dependencies for video salient object detection,'' \emph{IEEE TIP}, vol.~29, pp. 9017--9031, 2020.

\bibitem{r1_2}
X.~Zhou, Z.~Liu, C.~Gong, and W.~Liu, ``Improving video saliency detection via localized estimation and spatiotemporal refinement,'' \emph{IEEE Transactions on Multimedia}, vol.~20, no.~11, pp. 2993--3007, 2018.

\bibitem{r1_3}
T.~Huang, X.~Ben, C.~Gong, B.~Zhang, R.~Yan, and Q.~Wu, ``Enhanced spatial-temporal salience for cross-view gait recognition,'' \emph{IEEE Transactions on Circuits and Systems for Video Technology}, vol.~32, no.~10, pp. 6967--6980, 2022.

\bibitem{wang2021salient}
W.~Wang, Q.~Lai, H.~Fu, J.~Shen, H.~Ling, and R.~Yang, ``Salient object detection in the deep learning era: An in-depth survey,'' \emph{IEEE Transactions on Pattern Analysis and Machine Intelligence}, vol.~44, no.~6, pp. 3239--3259, 2021.

\bibitem{borji2019salient}
A.~Borji, M.-M. Cheng, Q.~Hou, H.~Jiang, and J.~Li, ``Salient object detection: A survey,'' \emph{Computational visual media}, vol.~5, pp. 117--150, 2019.

\bibitem{hou2017deeply}
Q.~Hou, M.-M. Cheng, X.~Hu, A.~Borji, Z.~Tu, and P.~H. Torr, ``Deeply supervised salient object detection with short connections,'' in \emph{CVPR}, 2017, pp. 3203--3212.

\bibitem{liu2019simple}
J.-J. Liu, Q.~Hou, M.-M. Cheng, J.~Feng, and J.~Jiang, ``A simple pooling-based design for real-time salient object detection,'' in \emph{CVPR}, 2019, pp. 3917--3926.

\bibitem{zhang2017amulet}
P.~Zhang, D.~Wang, H.~Lu, H.~Wang, and X.~Ruan, ``Amulet: Aggregating multi-level convolutional features for salient object detection,'' in \emph{ICCV}, 2017, pp. 202--211.

\bibitem{luo2017non}
Z.~Luo, A.~Mishra, A.~Achkar, J.~Eichel, S.~Li, and P.-M. Jodoin, ``Non-local deep features for salient object detection,'' in \emph{CVPR}, 2017, pp. 6609--6617.

\bibitem{wang2018detect}
T.~Wang, L.~Zhang, S.~Wang, H.~Lu, G.~Yang, X.~Ruan, and A.~Borji, ``Detect globally, refine locally: A novel approach to saliency detection,'' in \emph{CVPR}, 2018, pp. 3127--3135.

\bibitem{simonyan2014very}
K.~Simonyan and A.~Zisserman, ``Very deep convolutional networks for large-scale image recognition,'' in \emph{ICLR}, 2015.

\bibitem{Wang_2019_CVPR}
W.~Wang, J.~Shen, M.-M. Cheng, and L.~Shao, ``An iterative and cooperative top-down and bottom-up inference network for salient object detection,'' in \emph{Proceedings of the IEEE/CVF Conference on Computer Vision and Pattern Recognition (CVPR)}, June 2019.

\bibitem{wang2019inferring}
W.~Wang, J.~Shen, X.~Dong, A.~Borji, and R.~Yang, ``Inferring salient objects from human fixations,'' \emph{IEEE transactions on pattern analysis and machine intelligence}, vol.~42, no.~8, pp. 1913--1927, 2019.

\bibitem{xu2021locate}
B.~Xu, H.~Liang, R.~Liang, and P.~Chen, ``Locate globally, segment locally: A progressive architecture with knowledge review network for salient object detection,'' in \emph{Proceedings of the AAAI Conference on Artificial Intelligence}, vol.~35, no.~4, 2021, pp. 3004--3012.

\bibitem{zhao2019egnet}
J.-X. Zhao, J.-J. Liu, D.-P. Fan, Y.~Cao, J.~Yang, and M.-M. Cheng, ``Egnet: Edge guidance network for salient object detection,'' in \emph{ICCV}, 2019, pp. 8779--8788.

\bibitem{wang2018non}
X.~Wang, R.~Girshick, A.~Gupta, and K.~He, ``Non-local neural networks,'' in \emph{CVPR}, 2018, pp. 7794--7803.

\bibitem{bello2019attention}
I.~Bello, B.~Zoph, A.~Vaswani, J.~Shlens, and Q.~V. Le, ``Attention augmented convolutional networks,'' in \emph{ICCV}, 2019, pp. 3286--3295.

\bibitem{hu2019local}
H.~Hu, Z.~Zhang, Z.~Xie, and S.~Lin, ``Local relation networks for image recognition,'' in \emph{ICCV}, 2019, pp. 3464--3473.

\bibitem{zhao2020exploring}
H.~Zhao, J.~Jia, and V.~Koltun, ``Exploring self-attention for image recognition,'' in \emph{CVPR}, 2020, pp. 10\,076--10\,085.

\bibitem{ren2022shunted}
S.~Ren, D.~Zhou, S.~He, J.~Feng, and X.~Wang, ``Shunted self-attention via multi-scale token aggregation,'' in \emph{Proceedings of the IEEE/CVF Conference on Computer Vision and Pattern Recognition}, 2022, pp. 10\,853--10\,862.

\bibitem{ren2023tinymim}
S.~Ren, F.~Wei, Z.~Zhang, and H.~Hu, ``Tinymim: An empirical study of distilling mim pre-trained models,'' in \emph{Proceedings of the IEEE/CVF Conference on Computer Vision and Pattern Recognition}, 2023, pp. 3687--3697.

\bibitem{ren2023sg}
S.~Ren, X.~Yang, S.~Liu, and X.~Wang, ``Sg-former: Self-guided transformer with evolving token reallocation,'' in \emph{Proceedings of the IEEE/CVF International Conference on Computer Vision}, 2023, pp. 6003--6014.

\bibitem{wang2020axial}
H.~Wang, Y.~Zhu, B.~Green, H.~Adam, A.~Yuille, and L.-C. Chen, ``Axial-deeplab: Stand-alone axial-attention for panoptic segmentation,'' in \emph{ECCV}.\hskip 1em plus 0.5em minus 0.4em\relax Springer, 2020, pp. 108--126.

\bibitem{carion2020end}
N.~Carion, F.~Massa, G.~Synnaeve, N.~Usunier, A.~Kirillov, and S.~Zagoruyko, ``End-to-end object detection with transformers,'' in \emph{ECCV}.\hskip 1em plus 0.5em minus 0.4em\relax Springer, 2020, pp. 213--229.

\bibitem{zeng2020learning}
Y.~Zeng, J.~Fu, and H.~Chao, ``Learning joint spatial-temporal transformations for video inpainting,'' in \emph{ECCV}.\hskip 1em plus 0.5em minus 0.4em\relax Springer, 2020, pp. 528--543.

\bibitem{zheng2020rethinking}
S.~Zheng, J.~Lu, H.~Zhao, X.~Zhu, Z.~Luo, Y.~Wang, Y.~Fu, J.~Feng, T.~Xiang, P.~H. Torr \emph{et~al.}, ``Rethinking semantic segmentation from a sequence-to-sequence perspective with transformers,'' \emph{arXiv preprint arXiv:2012.15840}, 2020.

\bibitem{he2016deep}
K.~He, X.~Zhang, S.~Ren, and J.~Sun, ``Deep residual learning for image recognition,'' in \emph{CVPR}, 2016, pp. 770--778.

\bibitem{wang2019salient}
W.~Wang, S.~Zhao, J.~Shen, S.~C. Hoi, and A.~Borji, ``Salient object detection with pyramid attention and salient edges,'' in \emph{CVPR}, 2019, pp. 1448--1457.

\bibitem{wu2019stacked}
Z.~Wu, L.~Su, and Q.~Huang, ``Stacked cross refinement network for edge-aware salient object detection,'' in \emph{ICCV}, 2019, pp. 7264--7273.

\bibitem{chen2018reverse}
S.~Chen, X.~Tan, B.~Wang, and X.~Hu, ``Reverse attention for salient object detection,'' in \emph{ECCV}, 2018, pp. 234--250.

\bibitem{Wei_2020_CVPR}
J.~Wei, S.~Wang, Z.~Wu, C.~Su, Q.~Huang, and Q.~Tian, ``Label decoupling framework for salient object detection,'' in \emph{CVPR}, June 2020.

\bibitem{Qin_2019_CVPR}
X.~Qin, Z.~Zhang, C.~Huang, C.~Gao, M.~Dehghan, and M.~Jagersand, ``Basnet: Boundary-aware salient object detection,'' in \emph{CVPR}, June 2019.

\bibitem{Pang_2020_CVPR}
Y.~Pang, X.~Zhao, L.~Zhang, and H.~Lu, ``Multi-scale interactive network for salient object detection,'' in \emph{CVPR}, June 2020.

\bibitem{yang2013saliency}
C.~Yang, L.~Zhang, H.~Lu, X.~Ruan, and M.-H. Yang, ``Saliency detection via graph-based manifold ranking,'' in \emph{CVPR}.\hskip 1em plus 0.5em minus 0.4em\relax IEEE, 2013, pp. 3166--3173.

\bibitem{LiYu15}
G.~Li and Y.~Yu, ``Visual saliency based on multiscale deep features,'' in \emph{CVPR}, June 2015, pp. 5455--5463.

\bibitem{shi2015hierarchical}
J.~Shi, Q.~Yan, L.~Xu, and J.~Jia, ``Hierarchical image saliency detection on extended cssd,'' \emph{IEEE TPAMI}, vol.~38, no.~4, pp. 717--729, 2015.

\bibitem{wang2017}
L.~Wang, H.~Lu, Y.~Wang, M.~Feng, D.~Wang, B.~Yin, and X.~Ruan, ``Learning to detect salient objects with image-level supervision,'' in \emph{CVPR}, 2017.

\bibitem{li2014secrets}
Y.~Li, X.~Hou, C.~Koch, J.~M. Rehg, and A.~L. Yuille, ``The secrets of salient object segmentation,'' in \emph{CVPR}, 2014, pp. 280--287.

\bibitem{fan2017structure}
D.-P. Fan, M.-M. Cheng, Y.~Liu, T.~Li, and A.~Borji, ``Structure-measure: A new way to evaluate foreground maps,'' in \emph{ICCV}, 2017, pp. 4548--4557.

\bibitem{UCF}
P.~Zhang, D.~Wang, H.~Lu, H.~Wang, and B.~Yin, ``Learning uncertain convolutional features for accurate saliency detection,'' in \emph{ICCV}, Oct 2017.

\bibitem{Amulet}
P.~Zhang, D.~Wang, H.~Lu, H.~Wang, and X.~Ruan, ``Amulet: Aggregating multi-level convolutional features for salient object detection,'' in \emph{ICCV}, Oct 2017.

\bibitem{chen2018eccv}
S.~Chen, X.~Tan, B.~Wang, and X.~Hu, ``Reverse attention for salient object detection,'' in \emph{European Conference on Computer Vision}, 2018.

\bibitem{chang2018pyramid}
J.-R. Chang and Y.-S. Chen, ``Pyramid stereo matching network,'' in \emph{Proceedings of the IEEE Conference on Computer Vision and Pattern Recognition}, 2018, pp. 5410--5418.

\bibitem{Wu_2019_CVPR}
Z.~Wu, L.~Su, and Q.~Huang, ``Cascaded partial decoder for fast and accurate salient object detection,'' in \emph{CVPR}, June 2019.

\bibitem{9010954}
Z.~{Wu}, L.~{Su}, and Q.~{Huang}, ``Stacked cross refinement network for edge-aware salient object detection,'' in \emph{ICCV}, 2019, pp. 7263--7272.

\bibitem{hu2020sac}
X.~Hu, C.-W. Fu, L.~Zhu, T.~Wang, and P.-A. Heng, ``Sac-net: Spatial attenuation context for salient object detection,'' \emph{IEEE Transactions on Circuits and Systems for Video Technology}, 2020, to appear.

\bibitem{Liu21PamiPoolNet}
J.-J. Liu, Q.~Hou, Z.-A. Liu, and M.-M. Cheng, ``Poolnet+: Exploring the potential of pooling for salient object detection,'' \emph{IEEE TPAMI}, vol.~45, no.~1, pp. 887--904, 2023.

\bibitem{Ke_2022_WACV}
Y.~K. Yun and T.~Tsubono, ``Recursive contour-saliency blending network for accurate salient object detection,'' in \emph{Proceedings of the IEEE/CVF Winter Conference on Applications of Computer Vision (WACV)}, January 2022, pp. 2940--2950.

\bibitem{DBLP:journals/corr/abs-2104-12099}
\BIBentryALTinterwordspacing
N.~Liu, N.~Zhang, K.~Wan, J.~Han, and L.~Shao, ``Visual saliency transformer,'' \emph{CoRR}, vol. abs/2104.12099, 2021. [Online]. Available: \url{https://arxiv.org/abs/2104.12099}
\BIBentrySTDinterwordspacing

\bibitem{YAO2023342}
\BIBentryALTinterwordspacing
C.~Yao, L.~Feng, Y.~Kong, L.~Xiao, and T.~Chen, ``Transformers and cnns fusion network for salient object detection,'' \emph{Neurocomputing}, vol. 520, pp. 342--355, 2023. [Online]. Available: \url{https://www.sciencedirect.com/science/article/pii/S0925231222013704}
\BIBentrySTDinterwordspacing

\end{thebibliography}

\begin{IEEEbiography}[{\includegraphics[width=1in,height=1.25in,clip,keepaspectratio]{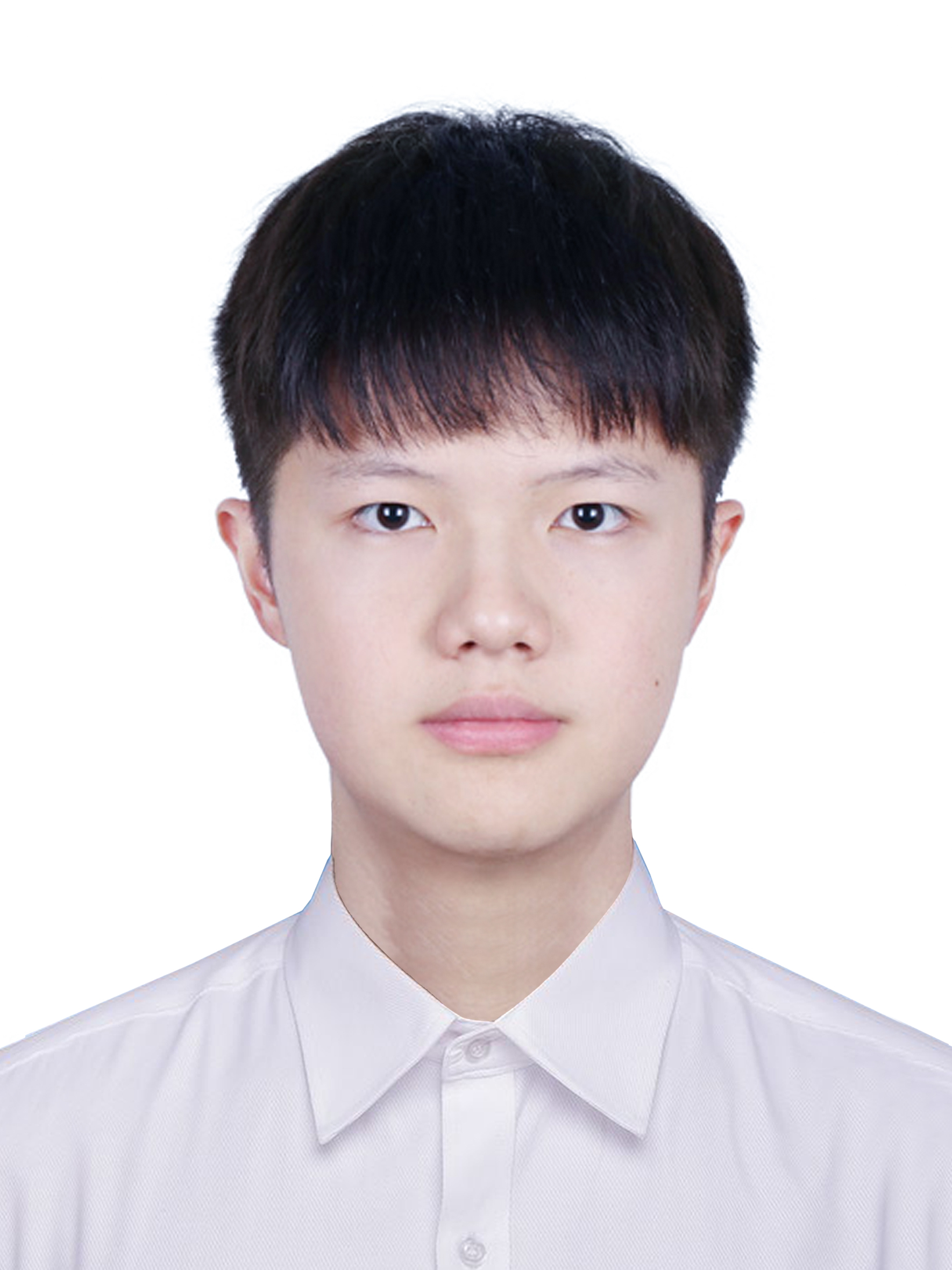}}]{Sucheng Ren} is a research assistant in the School of Computing and Information Systems, Singapore Management University. He received the M.Sc. degree and B. Eng. degree from South China University of Technology. His research interests include computer vision, image processing, and deep learning.
\end{IEEEbiography}

\begin{IEEEbiography}[{\includegraphics[width=1in,height=1.25in,clip,keepaspectratio]{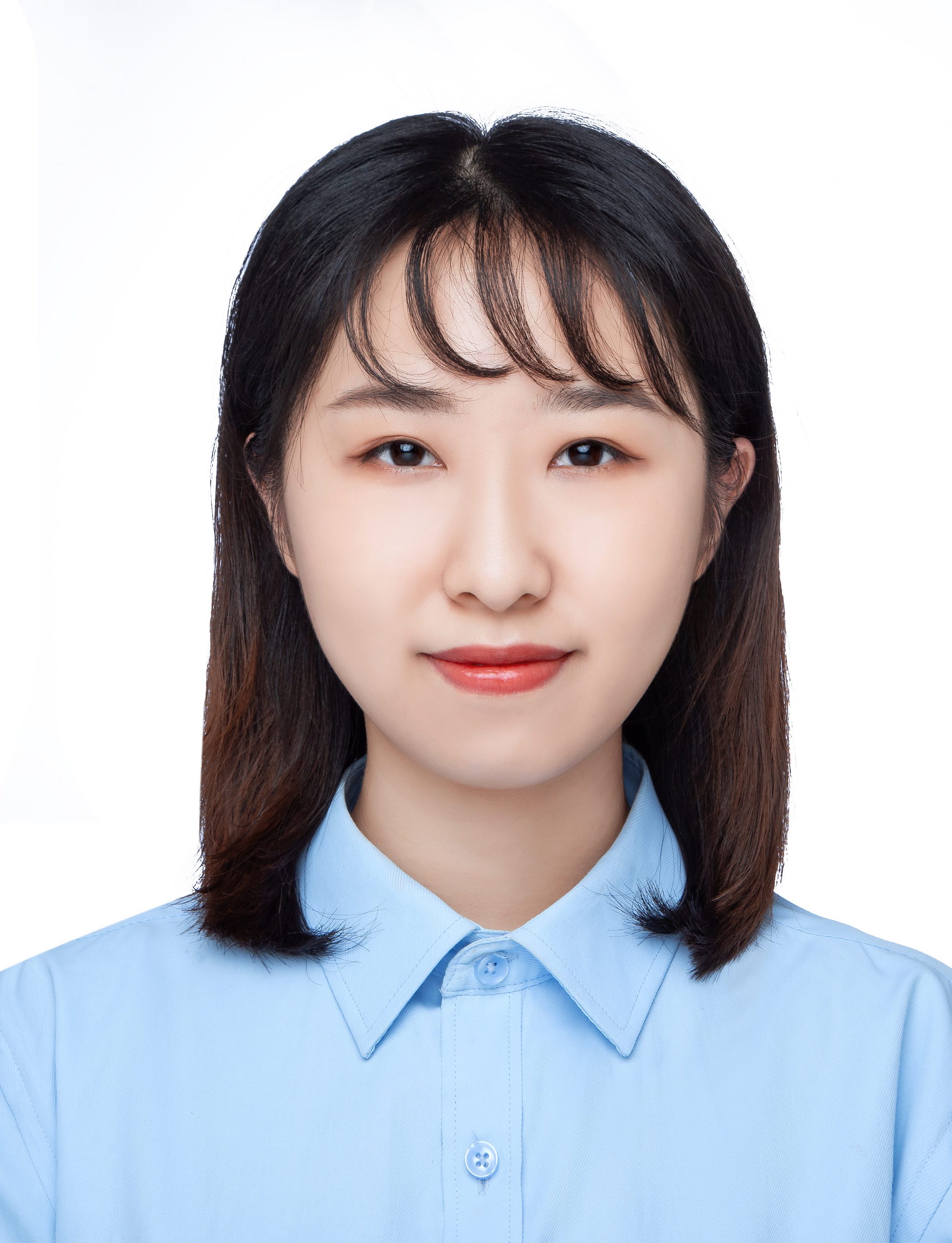}}]{Nanxuan Zhao} is a research scientist in Adobe Research. She was a Postdoctoral Researcher in the Chinese University of Hong Kong. She was a Visiting Scholar of Harvard University. She obtained her B.Sc. degree from South China University of Technology and her Ph.D. degree from City University of Hong Kong. Her research interests include computer graphics, computer vision, and human-computer interaction.
\end{IEEEbiography}

\begin{IEEEbiography}[{\includegraphics[width=1in,height=1.25in,clip,keepaspectratio]{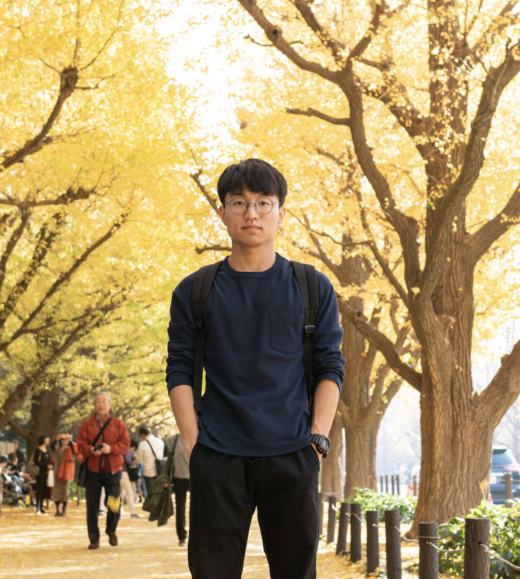}}]{Qiang Wen} was a master student in the School of Computer Science and Engineering, South China University of Technology. Before that, he received the B. Eng. degree from the School of Information Science and Engineering, Central South University in 2018. His research interests include computer vision, image processing and deep learning.

\end{IEEEbiography}

\begin{IEEEbiography}[{\includegraphics[width=1in,height=1.25in,clip,keepaspectratio]{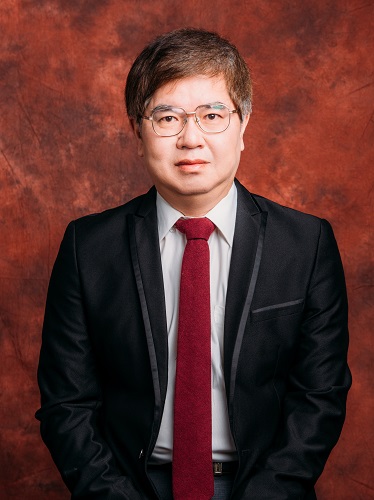}}]{Guoqiang Han}
	received the B.Sc. degree from the Zhejiang University, Hangzhou, China, in 1982, and the M.Sc. and Ph.D. degrees from the Sun Yat-sen University, Guangzhou, China, in 1985 and 1988, respectively.
	He is a Professor with the School of Computer Science and Engineering, South China University of Technology, Guangzhou. He was the dean of the
	School of Computer Science and Engineering. He has published over 100 research papers. His current research interests include multimedia, computational intelligence, machine learning, and computer graphics.
\end{IEEEbiography}

\begin{IEEEbiography}[{\includegraphics[width=1in,height=1.25in,clip,keepaspectratio]{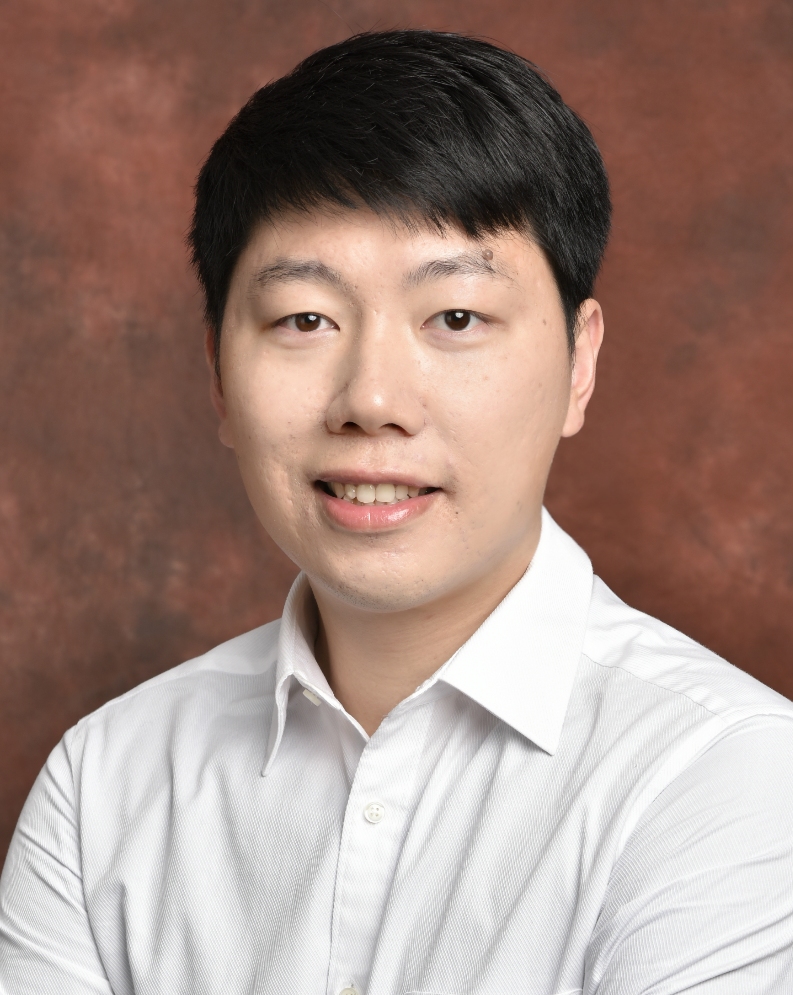}}]{Shengfeng He (Senior Member, IEEE)} is an associate professor in the School of Computing and Information Systems, Singapore Management University. He was on the faculty of the South China University of Technology, from 2016 to 2022. He obtained B.Sc. and M.Sc. degrees from Macau University of Science and Technology in 2009 and 2011 respectively, and a Ph.D. degree from City University of Hong Kong in 2015. His research interests include computer vision and generative models. He is a senior member of IEEE and CCF. He serves as the lead guest editor of the IJCV, the associate editor of IEEE TNNLS, IEEE TCSVT, Visual Intelligence, and Neurocomputing. He also serves as the area chair/senior program committee of ICML, AAAI, IJCAI, and BMVC.
\end{IEEEbiography}
\end{document}